%% file: AFNO.tex
\documentclass[twoside]{article}

\usepackage[accepted]{aistats2024}
\usepackage[utf8]{inputenc} % allow utf-8 input
\usepackage[T1]{fontenc}    % use 8-bit T1 fonts
\usepackage{hyperref}       % hyperlinks
\usepackage{url}            % simple URL typesetting
\usepackage{booktabs}       % professional-quality tables
\usepackage{amsfonts}       % blackboard math symbols
\usepackage{nicefrac}       % compact symbols for 1/2, etc.
\usepackage{microtype}      % microtypography
\usepackage{xcolor}         % colors

\usepackage{microtype}
\usepackage{graphicx}
\usepackage{booktabs} % for professional tables

\usepackage{microtype}
\usepackage{graphicx}
\usepackage{booktabs} % for professional tables

\usepackage{natbib}
\usepackage{amssymb, amsmath,amsthm}
\usepackage{amsfonts}
\usepackage{algorithm}
\usepackage{algorithmic}
\usepackage{paralist}
\usepackage{multirow}
\usepackage{times}
\usepackage{wrapfig}
\usepackage{url,enumerate}
\usepackage{color,xcolor}
\usepackage{makeidx}  % allows for indexgeneration
\usepackage{amsmath,amssymb}
\usepackage{mathtools}
\usepackage[small, compact]{titlesec}
\usepackage{xspace}
\usepackage{epstopdf}
\usepackage{cite}
\usepackage{mathrsfs}
\usepackage{times}
\usepackage{enumerate}
\usepackage{color}
\usepackage{graphicx,epsfig}
\usepackage{amsmath,amssymb,xspace}
\usepackage{url}
\usepackage{hyperref}
\usepackage{bm}
\usepackage{bbm}
\usepackage{upgreek}
\usepackage{cleveref}
\usepackage{multirow}
\usepackage{ulem}
\usepackage{cancel}
\usepackage{subcaption}
\usepackage{dsfont}
\usepackage{hyperref}

\newcommand{\ours}{MRA-FNO\xspace}

\newcommand{\cmt}[1]{}
\newcommand{\eg}{{\textit{e.g.},}\xspace}
\newcommand{\ie}{{\textit{i.e.},}\xspace}
\newcommand{\etc}{{\textit{etc}.}\xspace}

\input{emacscomm.tex}

\begin{document}

\runningauthor{Shibo Li, Xin Yu, Wei Xing, Robert M. Kirby, Akil Narayan, and Shandian Zhe}
\twocolumn[

\aistatstitle{Multi-Resolution Active Learning of Fourier Neural Operators}

\aistatsauthor{ Shibo Li$^1$  \And Xin Yu$^1$ \And Wei Xing$^{4}$ \And Robert M. Kirby$^{1,2}$ \And Akil Narayan$^{2,3}$  \And Shandian Zhe$^{1*}$}

\aistatsaddress{ $^1$Kahlert School of Computing, University of Utah \\
	$^2$Scientific Computing and Imaging (SCI) Institute, University of Utah\\
	$^3$Department
	of Mathematics, University of Utah\\
	%Salt Lake City, UT 84112\\
	$^4$School of Mathematics and Statistics, University of Sheffield\\
	\texttt{\{shibo, xiny, kirby, zhe\}@cs.utah.edu, akil@sci.utah.edu, w.xing@sheffield.ac.uk}
} ]

\input{./abstract}
\input{./intro}
\input{./method}
\input{./related}
\input{./exp-zhe}

\input{./conclusion}

\bibliographystyle{apalike}
\bibliography{AFNO}
\newpage
\appendix

\input{suppl-cr}

\end{document}

%% file: emacscomm.tex
\newcommand{\diag}{{\rm diag}}

\newcommand{\brho}{\boldsymbol{\rho}}
\renewcommand{\d}{{\rm d}}  % for derivatives
\newcommand{\e}{{\bf e}}
\newcommand{\f}{{\bf f}}
\newcommand{\g}{{\bf g}}
\newcommand{\h}{{\bf h}}

\newcommand{\wy}{{\widehat{\y}}}
\newcommand{\wlam}{{\widehat{\lambda}}}

\renewcommand{\v}{{\bf v}}

\newcommand{\x}{{\bf x}}
\newcommand{\y}{{\bf y}}

\newcommand{\B}{{\bf B}}

\newcommand{\Dcal}{\mathcal{D}}

\newcommand{\Pcal}{{\mathcal{P}}}
\newcommand{\Hcal}{{\mathcal{H}}}

\newcommand{\I}{{\bf I}}

\newcommand{\Ocal}{{\mathcal{O}}}
\newcommand{\Fcal}{{\mathcal{F}}}
\newcommand{\N}{\mathcal{N}}  % for normal density

\newcommand{\Wcal}{{\mathcal{W}}}

\newcommand{\Ycal}{{\mathcal{Y}}}

\newcommand{\bLambda}{\mathbf{\Lambda}}

\newcommand{\bmu}{\boldsymbol{\mu}}

\newcommand{\ben}{\begin{enumerate}}
\newcommand{\een}{\end{enumerate}}

\newcommand{\argmin}{\operatornamewithlimits{argmin}}
\newcommand{\argmax}{\operatornamewithlimits{argmax}}

\newcommand{\EE}{\mathbb{E}}

%% file: abstract.tex
%Operator learning --> muti-reoslution --> ....
\begin{abstract}
	Fourier Neural Operator (FNO) is a popular operator learning framework. It not only achieves the state-of-the-art performance in many tasks, but also is efficient in training and prediction. However, collecting training data for the FNO can be a costly bottleneck in practice, because it often demands expensive physical simulations. To overcome this problem, we propose  Multi-Resolution Active learning of FNO (\ours), which can dynamically select the input functions and resolutions to lower the data cost as much as possible while optimizing the learning efficiency. Specifically, we propose a probabilistic multi-resolution FNO and use ensemble Monte-Carlo to develop an effective posterior inference algorithm. To conduct active learning, we maximize a utility-cost ratio as the acquisition function to acquire new examples and resolutions at each step. We use moment matching and the matrix determinant lemma to enable tractable, efficient utility computation. Furthermore, we develop a cost annealing framework to avoid over-penalizing high-resolution queries at the early stage. The over-penalization is severe when the cost difference  is significant between the resolutions, which renders active learning often stuck at low-resolution queries and inferior performance. Our method overcomes this problem and  applies to general multi-fidelity active learning and optimization problems. We have shown  the advantage of our method in several benchmark operator learning tasks. The code is available at \url{https://github.com/shib0li/MRA-FNO}.
\end{abstract}

%% file: intro.tex
%operator learning background, FNO --> multi-resolution training, reduce the cost --> active learning our contributions --> result 
\section{INTRODUCTION}
%applications  & background --> 
Operator learning is emerging as an important topic in scientific machine learning. It intends to estimate function-to-function mappings and can serve as a useful surrogate model for many physical simulation related applications, such as weather forecast~\citep{pathak2022fourcastnet}, control~\citep{bhan2023operator}, engineering design~\citep{liu2023deepoheat} and inverse problems~\citep{kaltenbach2022semi}. One representative  approach is the Fourier neural operator (FNO)~\citep{li2020fourier}, which uses fast Fourier transform (FFT) and convolution theorem to implement global linear transforms in the functional space. The FNO not only shows state-of-the-art performance in many tasks, but also is highly efficient in training and prediction. 

%Operator learning has been emerging as an important topic in \textit{AI for Science}. It intends to estimate function-to-function mappings and can serve as a useful surrogate model for many physical simulation related applications, such as weather forecast [XX], engineering design [XX] and inverse problems [XX]. One popular and representative  approach is Fourier neural operators (FNO), which uses fast Fourier transform (FFT) and the convolution theorem to fulfill a linear transform in functional spaces. The Fourier layers in FNO actually define a novel global spectral convolution operation.  FNO's not only have achieved the state-of-the-art prediction accuracy in many problems, but also are highly efficient in training and prediction [XX]. 

%we propose an active learning approach: (1) model (2) active learning, computational trick (3) annealing appraoch (4) experiment
Despite the advantages, collecting training data for the FNO can be a severe bottleneck in practice because it  often requires many physical simulations (\eg running numerical solvers), which is known to be computationally expensive.  To reduce the cost, one can consider leveraging multi-resolution data. The low-resolution data is cheap to obtain  --- typically computed with rough meshes --- but the provided output function samples are quite inaccurate (large bias). On the contrary,   high-resolution data offers accurate output function samples, yet is much more costly to generate from dense meshes. Although with substantial difference in quality, the low and high resolution examples share the same underlying physics and are strongly correlated. Hence, one can reasonably expect using multi-resolution data to well train the FNO while reducing the data cost. 
%to well train an FNO with multi-resolution data while reducing the data cost. 

However, blindly collecting examples at different resolutions is hardly optimal in both cost saving and learning efficiency.  To reduce the data cost to the greatest extent while optimizing the learning efficiency, we propose \ours, a novel multi-resolution active learning method, which can dynamically select the best input function and resolution each step, at which to generate new examples.
The major contributions of our work are summarized as follows. 
\begin{itemize}
	\item \textbf{Probabilistic Multi-Resolution FNO.} We first extend the FNO to integrate multi-resolution training data.  To capture the influence of the resolution choice on the predictive distribution, we append a resolution embedding to the samples of the input function. After the FNO layers,  we create two branches: one generates the  prediction mean of the target function and the other the variance. In this way, the prediction is up to not only the input function samples but also the resolution choice.  We then use Monte-Carlo ensemble learning~\citep{lakshminarayanan2017simple} to fulfill effective uncertainty quantification, which is critical for utility evaluation and active learning.
	%applies 	a feed-forward network channel-wisely to generate the predictive mean of the target function, and the other uses convolutional and feed-forward layers to generate the predictive variance. In this way, both the predictive mean and variance are up to not only the input function samples but also the resolution choice. 
%	\item\textbf{Probabilistic Inference.} To enable effective uncertainty quantification, which is critical for utility evaluation in active learning, we construct a discrete 
%	 we use the ensemble Monte-Carlo to develop a posterior inference algorithm. An ensemble of model parameters are estimated from independent random initializations and stochastic training. We then construct a discrete posterior approximation of the model parameters, with which we obtain a closed-form\cmt{Gaussian mixture } predictive distribution. 
	%annealing 
	\item \textbf{Active Learning.} %To optimize the learning efficiency while reducing the data cost as much as possible, we use a similar strategy to the state-of-the-art multi-fidelity active learning and Bayesian optimization methods~\citep{li2022deep,takeno2020multi,li2020multi}. That is,  we 
	% maximize the utility-cost ratio to determine the training input and resolution at each step, where the utility is measured by mutual information. However, we found two severe challenges. The first challenge is that the computation of the utility function is analytically intractable and costly.   
	To optimize the learning efficiency while reducing the data cost as much as possible, we maximize the utility-cost ratio to select the best training input and resolution at each step, where the utility is measured by mutual information. The strategy is similar to the state-of-the-art multi-fidelity active learning and Bayesian optimization methods~\citep{li2022deep,takeno2020multi,li2020multi}, but there are two severe challenges. The first challenge is that the computation of the utility function is  analytically intractable and costly.  
	 We use moment matching to approximate the posterior predictive distribution as a multi-variate Gaussian. We then leverage the structure of the covariance matrix, and apply the matrix determinant lemma to achieve efficient, closed-form mutual information calculation. The second challenge  is that, directly maximizing the utility-cost ratio as in previous methods,  tends to trap the active learning at low-resolution queries and inferior performance. This is due to that when the data is few (at the early stage), the mutual information measurement for examples at different resolutions is close. High-resolution examples are thereby over-penalized by the large cost.\cmt{The value of high-resolution examples is actually under-estimated or (in other words) over-penalized by the large cost.}
	  We propose a cost annealing framework, which initializes the same cost for every resolution. The cost for each resolution is scheduled to gradually converge to the true cost along with data accumulation. When the data is enough and  mutual information can reflect the true potential of each example, our active learning returns to maximizing the benefit-cost ratio.	In this way, our method can flexibly incorporate high-resolution examples at the early stage to ensure continuous improvement. Our framework applies to general multi-fidelity learning and optimization problems.
	%annelaning and computational trick %We propose 
	\item \textbf{Experimental Results.} We evaluated \ours with four benchmark operator learning tasks, based on Burger's, Darcy flow, nonlinear diffusion and Navier-Stoke equations. On fixed training datasets, our multi-resolution FNO shows better or very close prediction error as compared to the standard FNO.  Both the prediction accuracy and test log likelihood are much higher than applying other popular Bayesian inference methods, including Monte-Carlo (MC) dropout, stochastic gradient Langevin dynamics and variational inference. It shows our  ensemble inference provides much better uncertainty quantification. During the course of each active learning experiment, \ours consistently achieves much better prediction accuracy with the same accumulated data cost, as compared with random queries,  core-set active learning, and our framework using dropout inference. 
	
%	First, on fixed training datasets, our multi-resolution FNO model exhibits much better prediction accuracy than the standard FNO that simply integrates all examples with different resolutions. Next, our ensemble Monte-Carlo posterior inference results in much better prediction accuracy and uncertainty quantification than other popular Bayesian inference methods, including Monte-Carlo dropout, SGLD and variational inference. Third, in five active learning experiments, \ours consistently shows much better prediction accuracy with the same accumulated data cost, as compared with random queries,  core-set active learning strategies, and using our framework but with Monte-Carlo dropout inference. 
\end{itemize}

%% file: method.tex
%\vspace{-0.1in}
%\vspace{-0.1in}
%hyperparameter tuning ---> BO formulation, MES principle
\section{BACKGROUND}
%\vspace{-0.1in}
%need to: highlight complexity of M and large scale of Dtr
%\subsection{Fourier Neural Operators}
\textbf{Operator Learning.} Suppose our goal is to learn a function-to-function mapping $\psi: \Hcal \rightarrow \Ycal$, where $\Hcal$ and $\Ycal$ are two function spaces, \eg Banach spaces. The training dataset comprises pairs of discretized input and output functions, $\Dcal = \{(\f_n, \y_n)\}_{n=1}^N$, where each $\f_n$ are samples of a function $f_n \in \Hcal$, and $\y_n$ are samples of $\psi[f_n] \in \Ycal$.  All the input and output functions are discretized (sampled) at a set of evenly-spaced locations, \eg  a $64 \times 64$ mesh in the 2D spatial domain $[0, 1] \times[0, 1]$. 

\textbf{Fourier Neural Operators (FNO).} Given a discretized input function $\f$, the FNO first applies a feed-forward network (FFN) over each element of $\f$ and the associated sampling location to lift the input to a higher-dimensional channel space. Then a Fourier layer is used  perform a linear transform and nonlinear activation in the functional space, 
\begin{align}
	v(\x) \leftarrow \sigma\left(\Wcal v(\x) + \int \kappa(\x - \x')v(\x') \d \x' \right) \notag 
\end{align}
where $v(\x)$ in the R.H.S is the input function to the Fourier layer and in the L.H.S the output function, $\kappa(\cdot)$ is the integration kernel and $\sigma(\cdot)$ is the activation. Based on the convolution theorem $ \int \kappa(\x - \x')v(\x') \d \x' = \Fcal^{-1}\left[\Fcal[\kappa]\cdot\Fcal[v]\right](\x)$ where $\Fcal$ and $\Fcal^{-1}$ are the Fourier and inverse Fourier transforms, respectively, the Fourier layer  performs fast Fourier transform (FFT) over $v$, multiplies it with the discretized kernel in the frequency domain,  and then performs inverse FFT. The local linear transform, $\Wcal v(\x)$, is performed by standard convolution (as in convolution nets). Due to the usage of FFT, the computation of the Fourier layer is highly efficient. After several Fourier layers, another FFN is applied channel-wisely to project back and make the final prediction. The training is typically done by minimizing an $L_2$ loss, $\Theta^* = \argmin_{\Theta}\frac{1}{N}\sum_{n=1}^N\|\g_n - \psi_{\text{FNO}}(\f_n;\Theta)\| $, \cmt{
\begin{align}
	\Theta^* = \argmin_{\Theta}\frac{1}{N}\sum_{n=1}^N\|\g_n - \psi_{\text{FNO}}(\f_n;\Theta)\| \label{eq:fno-loss}
\end{align}
}
where $\Theta$ are the model parameters,  including the discretized kernel in the frequency domain, standard convolution parameters in each Fourier layer, and the parameters of the FNN's for channel lifting and projection. 
%$(f_j, \x_j)$ where $\x_j$ is the sample location $j$ so as to lift the input function to a higher-dimensional space. 
%Given each training input $\f_n$, the Fourier neural operator (FNO) first lifts $\f_n$ to higher-dimensional channel space via applying a feed-forward network (FFN) over each pair $([\f_n]_j, \x_j)$ where $\x_j$ the $j$-the sample location. 
\cmt{
\begin{figure*}[h!]
	\centering
	\setlength\tabcolsep{0pt}
	\begin{tabular}[c]{c}
		\setcounter{subfigure}{0}
		\begin{subfigure}[t]{1.0\textwidth}
			\centering
			\includegraphics[width=\textwidth]{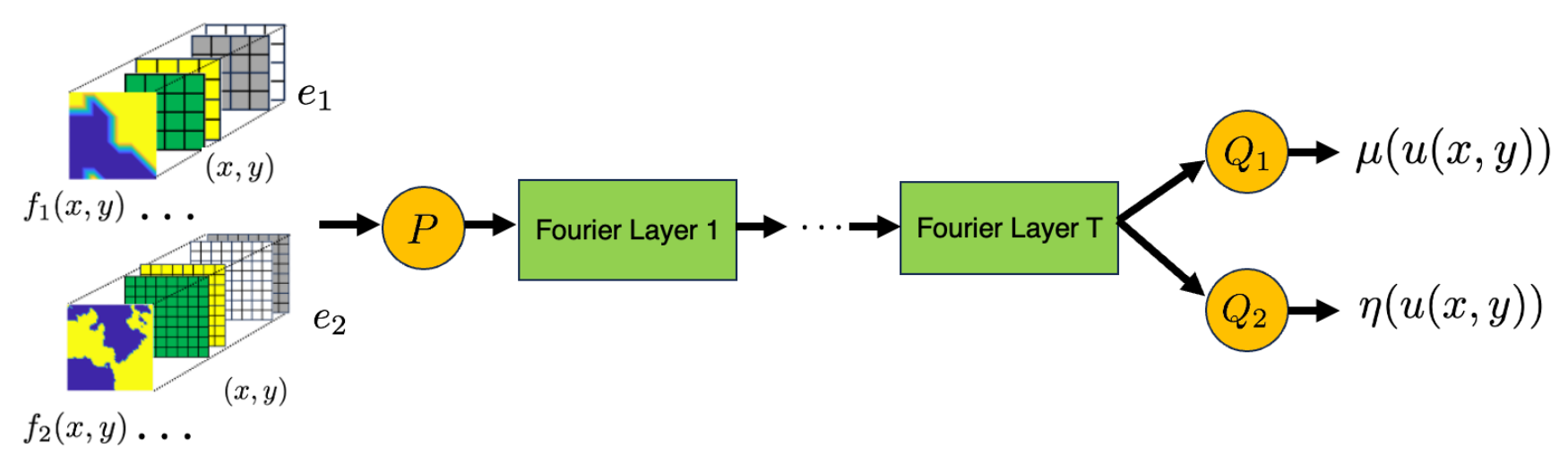}
		\end{subfigure}
	\end{tabular}
	%\vspace{-0.1in}
	\caption{\small Graphical representation of our probabilistic multi-resolution FNO. Here $P$ is the FFN that lifts the input function to higher-dimensional channel space, $Q_1$ is the FFN for channel-wise projection and producing the predictive mean, 
		and $Q_2$ is a convolution net plus another FFN to produce the predictive variance in the log space. } \label{fig:model}
\end{figure*}
}
%\section{Probabilistic Multi-Resolution FNO}\label{sect:model}
\section{PROBABILISTIC MULTI-RESOLUTION FNO}\label{sect:model}
%talk about the bottleneck, and multi-resoultion learning --> model design --> probabilistic inference (should be done in the morning < 12:00pm)
Despite the advantages of the FNO,  the training data collection can be a severe bottleneck  for  practical usage, because it typically requires many expensive physical simulations. To reduce the cost, we consider using multi-resolution data, which combines accurate yet expensive high-resolution examples with inaccurate (large bias) yet cheap-to-generate low-resolution examples. We then propose an active learning approach \cmt{that can dynamically select each training example and resolution according to the learning progress so as} to lower the data cost to the fullest extent while reaching a high learning efficiency. To this end, we first propose a probabilistic FNO that can effectively  integrate multi-resolution training examples and  perform posterior inference.

Specifically, suppose a multi-resolution dataset is given, $\Dcal = \{(\f_n, \g_n, r_n)\}_{n=1}^N$ where $r_n$ denotes the resolution of the $n$-th example. We have $R$ different resolutions in total ($1 \le r_n \le R$). For example, on a 2D spatial domain $[0, 1] \times [0, 1 ]$,  we might have two resolutions, $16 \times 16$ and $128 \times 128$.  To explicitly model the influence of the resolution choice on the prediction, we introduce an embedding  $\e_r$ to represent each resolution $r \in [1, R]$. {In our experiment, we set $\e_r$ to a one-hot encoding.  We have also tried  other embeddings, such as positional encodings~\citep{vaswani2017attention}.} The performance is close.  We apply an FFN to every element of $\f_n$, the corresponding sample location $\x_j$, and the embedding $\e_{r_n}$ to obtain a new representation $\widehat{\f}_n$,  where each
\begin{align}
[\widehat{\f}_n]_j = \text{FNN}([\f_n]_j, \x_j, \e_{r_n}).
\end{align}
Next, we use standard Fourier layers to perform successive linear and nonlinear transforms in the functional space. Denote by $\v_n$ the output (discretized) function. We then create two branches. One branch applies an FNN in each channel to project $\v_n$ back to the target dimension and output the prediction mean, $\bmu_\Theta(\f_n, \e_n)$ where $\Theta$ denote the model parameters. The other branch performs a standard convolution and then an FNN to output the prediction variance in the log domain, $\eta_\Theta(\f_n, \e_n)$. We then use a Gaussian likelihood to model the observed (discretized) output function, 
\begin{align}
	p(\g_n|\f_n, r_n) =  \N\left(\g_n | \bmu_\Theta(\f_n, \e_{r_n}), e^{\eta_\Theta(\f_n, \e_{r_n})}\cdot\I\right). \notag 
\end{align}
We can see that both the mean and variance are not only dependent on the input $\f_n$ but also up to the resolution choice $r_n$. In this way, our model can capture the influence of the resolution choice on the prediction distribution. %One can naturally expect a larger variance at low resolutions, reflecting more uncertainty. In addition, the prediction at high resolutions should be more confident and gives a smaller variance. 
Our model is illustrated in Appendix Fig. \ref{fig:model}.
%, especially on the bias toward the observations, which are modeled by the log predictive variance $\eta$. One can naturally expect the prediction should have a larger bias toward low-resolution examples and small bias toward high-resolution examples. Our model is illustrated in Fig. XX.  %Although the standard FNO is resolution free and we can directly use all the examples of different resolutions for training, the model essentially assumes the same prediction bias (or error) for different resolutions (see \eqref{eq:fno-loss}), which is misspecified. Our model is illustrated in Fig. XX. 

Next, we use Monte-Carlo ensemble learning~\citep{lakshminarayanan2017simple}\footnote{we do not introduce adversarial samples as in~\citep{lakshminarayanan2017simple}. We empirically found little help with such samples.} to fulfill effective posterior inference. Specifically, we randomly initialize the model parameters $\Theta$, and maximize the log likelihood to obtain one point estimate via stochastic mini-batch optimization, 
\begin{align}
	\Theta^* &= \argmax_\Theta \;\; \sum_{n=1}^N \log\left[\N\left(\g_n | \bmu_\Theta(\f_n, \e_n), e^{\eta_\Theta(\f_n, \e_n)}\I\right) \right]. \notag 
\end{align}
We independently repeat this procedure for $M$ times, and obtain an ensemble of the point estimates of the model parameters, $\{\Theta^*_1, \ldots, \Theta^*_M\}$. We then construct a discrete posterior approximation of the model parameters, 
%\begin{align}
	$p(\Theta|\Dcal) \approx \frac{1}{M} \sum_{m=1}^M \delta(\Theta - \Theta^*_m)$, 
%\end{align}
where $\delta(\cdot)$ is the Dirac delta measure. Given a test input function $\f$ and the resolution embedding $\e$, the predictive distribution of the output function is therefore a Gaussian mixture, 
\begin{align}
	&p(\y(\f, \e)|\Dcal) \notag \\
	&=\frac{1}{M}   \sum_{m=1}^M \N\left(\y|  \bmu_{\Theta^*_m}(\f, \e), \sigma^2_{\Theta^*_m}(\f, \e)\cdot \I\right). \label{eq:pred-dist}
\end{align}
where $\sigma^2_{\Theta^*_m}(\f, \e) = e^{\eta_{\Theta^*_m}(\f, \e)}$.

%\section{Multi-Resolution Active Learning} \label{sect:mral}
\section{MULTI-RESOLUTION ACTIVE LEARNING} \label{sect:mral}
%\vspace{-0.1in}
%general strategy, acquisition function def,  computation 
Now, we present our multi-resolution active learning algorithm.  To optimize the learning efficiency while lowering the data cost as much as possible, at each step, we maximize a utility-cost ratio (as the acquisition function) to determine the most valuable input function and its resolution, at which we query a new example. Specifically, we prepare a pool of candidate input functions $\Pcal$. Denote by $\lambda_r$ the cost of generating the output function at resolution $r \in [1, R]$. We have $\lambda_1 < \ldots <\lambda_R$. To measure the value of an example with input function $h \in \Pcal$ and resolution $r$, we consider two utility functions. The first one follows~\citep{li2022deep} and quantifies the information the example can bring to predict at the highest resolution $R$, 
\begin{align}
	u(h, r) = \mathbb{I}(\y(\h^r, \e_r), \y(\h^R, \e_R) |\Dcal),  \label{eq:u1}
\end{align}
where $\Dcal$ is the current training dataset, $\mathbb{I}(\cdot, \cdot)$ is the mutual information, $\h^r$ and $\h^R$ are function $h$ discretized at resolution $r$ and $R$, respectively, and $\e_r$ and $\e_R$ are the corresponding resolution embeddings.\cmt{, and $\y(\h^r, \e^r)$ and $\y(\h^d, \e^d)$ are the model predictions at  resolution $d$ and $r$, respectively; see \eqref{eq:pred-dist}.} The utility function \eqref{eq:u1} only considers how the example can improve the prediction for the same input function. To model its benefit in improving the prediction for other input functions, we follow~\citep{li2022batch} to consider a second utility  function $u(h, r) = \EE_{p(h')}[\mathbb{I}(\y(\h^r, \e_r), \y(\h'^R, \e_R)|\Dcal)]$, where $h' \in \Hcal$ and  $p(h')$ is a distribution over $\Hcal$. The expectation usually does not have a closed-form, and we therefore draw $A$ functions, $h'_1, \ldots, h'_A \sim p(h')$, and employ an Monte-Carlo approximation, 
\begin{align}
	\widehat{u}(h, r) = \frac{1}{A}\sum_{l=1}^A \mathbb{I}(\y(\h^r, \e_r), \y(\h_l'^R, \e_R) |\Dcal).  \label{eq:u2}
\end{align}
\cmt{
At each step, we maximize the utility-cost ratio to obtain the optimal input function and resolution, at which to generate a new training example, 
\begin{align}
	h^*, r^*  = \argmax_{h \in \Pcal, 1 \le r \le d} \frac{\beta(h, r) }{\lambda_{r}} \label{eq:acqn-1}
\end{align}
where $\beta(h, r)$ takes $u(h, r)$ in \eqref{eq:u1} or $\widehat{u}(h, r)$ in \eqref{eq:u2}. 
}
\subsection{Efficient Utility Computation}
The utility function in both \eqref{eq:u1} and \eqref{eq:u2} demands we compute the mutual information between a pair of predictions from our model. The computation is challenging in that (1) those predictions are typically high-dimensional (\eg a  $100 \times 100$ resolution corresponds to $10$K dimensional outputs), and (2) the mutual information is analytically intractable due to the Gaussian mixture predictive distribution in \eqref{eq:pred-dist}. To address this problem, we observe that for any two predictions $\y_1$ and $\y_2$, we have
\begin{align}
	&\mathbb{I}(\y_1, \y_2|\Dcal) \notag \\
	&= \mathbb{H}(\y_1|\Dcal) + \mathbb{H}(\y_2|\Dcal) - \mathbb{H}(\y_1, \y_2|\Dcal). \label{eq:MI}
\end{align}
Denote by $(\f_1, \e_1)$ the discretized input function and resolution embedding for $\y_1$ and by $(\f_2, \e_2)$ for $\y_2$.  We first use moment matching to approximate the predictive distributions of $\y_1$, $\y_2$ and $\wy = [\y_1; \y_2]$ as multi-variate Gaussian distributions, and we can thereby compute each entropy with a closed form. Specifically, let us first consider $\wy$. According to \eqref{eq:pred-dist}, we can derive that 
%\begin{align}
	$p(\wy|\Dcal) = \frac{1}{M} \sum_{m=1}^M \N(\wy | \brho_m, \bLambda_m)$, 
%\end{align}
where $\brho_m = [\bmu_{\Theta^*_m}(\f_1, \e_1); \bmu_{\Theta^*_m}(\f_2, \e_2)]$ and $\bLambda_m = \diag\left(\sigma^2_{\Theta^*_m}(\f_1, \e_1)\cdot\I, \sigma^2_{\Theta^*_m}(\f_2, \e_2)\cdot\I\right)$. The mean and covariance, \ie the first and second moments,  are given by
\begin{align}
  &\EE(\wy|\Dcal) = \frac{1}{M} \sum_{m=1}^M \brho_m, \notag \\
  & \text{cov}(\wy|\Dcal) = \frac{1}{M}\sum_{m=1}^M \left(\bLambda_m + \brho_m \brho_m^\top\right) - \EE(\wy|\Dcal)\EE(\wy|\Dcal)^\top.  \notag 
\end{align}
Via moment matching, we construct a multi-variate Gaussian approximation, $p(\wy|\Dcal) \approx \N(\wy| \EE(\wy|\Dcal), \text{cov}(\wy|\Dcal))$, which is the best approximation in the exponential family in the sense of Kullback Leibler divergence~\citep{bishop2006pattern}.  \cmt{Note that the moment matching minimizes the Kullback–Leibler divergence from the target distribution to the approximate distribution in the exponential family.}   Accordingly, the entropy can be computed with a closed-form, $\mathbb{H}(\wy) = \frac{1}{2}\log \text{det} \left[\text{cov}(\wy|\Dcal) \right] + \text{const}$.

However, since $\wy$ is high-dimensional, computing the log determinant of its huge covariance matrix is extremely expensive or even infeasible. To address this problem, we observe that
\begin{align}
	 &\text{cov}(\wy|\Dcal) =\bLambda +  \frac{1}{M}\sum_{m=1}^M  \brho_m \brho_m^\top - \EE(\wy|\Dcal)\EE(\wy|\Dcal)^\top
	  \notag \\
	 &=\bLambda + \frac{1}{M-1} \sum_{m=1}^M\left(\brho_m - \EE(\wy|\Dcal)\right)\left(\brho_m - \EE(\wy|\Dcal)\right)^\top \label{eq:result}
\end{align}
where $$\bLambda = \diag\left( \frac{1}{M}\sum_{m=1}^M \sigma^2_{\Theta^*_j}(\f_1, \e_1) \cdot \I, \frac{1}{M}\sum_{m=1}^M \sigma^2_{\Theta^*_j}(\f_2, \e_2)\cdot\I  \right) $$ is a diagonal matrix, and the second term in the R.H.S of \eqref{eq:result} is the empirical covariance matrix over $\{\brho_m\}$. We can further derive that
%\begin{align}
	$\text{cov}(\wy|\Dcal)  = \bLambda  + \B \B^\top$, 
%\end{align}
where $\B = \frac{1}{\sqrt{M-1}} [\brho_1 - \EE(\wy|\Dcal), \ldots, \brho_M - \EE(\wy|\Dcal)]$, which includes $M$ columns. We then use the matrix determinant lemma~\citep{harville1997matrix} to compute,  
\begin{align}
	&\log\text{det} \left[\text{cov}(\wy|\Dcal) \right] = \log\text{det}\left[ \bLambda  + \B \B^\top \right] \notag \\
	&= \log\text{det}[\bLambda] +  \log\text{det}[\I + \B^\top \bLambda^{-1} \B]. 
\end{align}
The first log determinant is over the diagonal matrix $\bLambda$, and the complexity is linear in the dimension of $\wy$. The second log determinant is computed over an $M \times M$ matrix. Since $M$ is the size of the ensemble and is very small (we take $M=5$ in our experiments), the computation is highly efficient. It is straightforward to use a similar approach to compute $\mathbb{H}(\y_1|\Dcal)$ and $\mathbb{H}(\y_2|\Dcal)$ in \eqref{eq:MI}. 

\subsection{Cost Annealing}
In practice,  directly maximizing the utility-cost ratio $\frac{u(h,r)}{\lambda_{r}}$ or $\frac{\widehat{u}(h,r)}{\lambda_{r}}$ (see \eqref{eq:u1} and \eqref{eq:u2}) tends to make the active learning stuck at low-resolution queries  and inferior performance, especially when the cost discrepancy is significant between the low and high resolutions. This is because at the early stage, the training data is few, and the mutual information\cmt{ (quantified based on our ensemble posterior inference)} does not differ much for candidates at different resolutions. In other words, the scales are close. Consequently, the high-resolution examples are over-penalized by the large cost, and the active learning keeps selecting low-resolution examples, which can severely hinder the model improvement. 

To overcome this problem, we propose a cost annealing method. We schedule a dynamic cost assignment for each resolution. Denote by $\wlam_r(t)$ the cost schedule for resolution $r$ at step $t$. For convenience, we normalize the true cost into $[0, 1]$, \ie  each $\lambda_r \in [0, 1]$ and $\sum_{r=1}^R \lambda_r = 1$. We set
\begin{align}
	\wlam_r(t) = \frac{\lambda_r}{1 + (R\lambda_r - 1)c(t)}, \label{eq:cost-schedule}
\end{align}
where $c(t)$ is a decaying function such that $c(0) = 1$ and $c(\infty) = 0$. For example, we can use 
\begin{align}
	c(t) = \exp(-\alpha t), \;\; \text{or}\;\; c(t) = 2(1 - s(\alpha t)), \label{eq:decay-func}
\end{align}
where $s(\cdot)$ is the sigmoid function and $\alpha$ controls the decay rate. We can see that all $\wlam_r(0) = \frac{1}{R}$ and $\lim\limits_{t \rightarrow \infty} \wlam_r(t) = \lambda_r$. To further enhance smoothness in annealing, we can re-normalize $\wlam_r$ at each step $t$. Note that this adjustment does not alter the convergence limit. We select the input and resolution by maximizing the acquisition function, $\frac{u(h,r)}{\wlam_{r}(t)}$ or $\frac{\widehat{u}(h,r)}{\wlam_{r}(t)}$. In this way, at the early stage when the data is few and the mutual information does not differ much, our method avoids over-penalizing high-resolution examples, and promote their queries to ensure continuous model improvement. With the accumulation of  data,  the mutual information is more and more capable of reflecting the true potential/value of new examples, the active learning returns to maximizing the ideal utility-cost ratio to select the input functions and resolutions. Our method is summarized in Algorithm \ref{alg:al}. 

\subsection{Algorithm Complexity}
%\textbf{Algorithm Complexity.} 
The time complexity of each active learning step is $\Ocal(|\Pcal|RM^2d)$ where $|\Pcal|$ is the size of the candidate pool, and $d$ is the output dimension at the highest resolution. The space complexity is $\Ocal(Md)$, which is to store the predictive distribution (for any input function) and the parameter estimates in the ensemble.  
\begin{figure}[t]
\begin{algorithm}[H]
	\small 
	\caption{\ours($M$, $\Pcal$, $T$, $\{\lambda_r\}_{r=1}^R$)}          
	\label{alg:al}                           
	\begin{algorithmic}[1]                    % enter the algorithmic environment
		\STATE Learn the probabilistic multi-resolution FNO from an initial dataset $\Dcal$ with the ensemble size $M$. 
		\FOR{$t=1 \ldots T$}
		\STATE Based on the cost schedule \eqref{eq:cost-schedule}, select the input function $h_t \in \Pcal$ and the resolution $r_t$ by  
		\begin{align}
			h_t, r_t  = \argmax_{h \in \Pcal, 1 \le r \le R} \frac{\beta(h, r) }{\wlam_{r}(t)} \notag 
		\end{align}
	where $\beta(h, r)$ is the utility function that can take \eqref{eq:u1} or \eqref{eq:u2}. 
		\STATE Query the output function at $h_t$ with resolution $r_t$ to obtain $\y_t$.
		\STATE Remove $h_t$ from $\Pcal$. 
		\STATE $\Dcal \leftarrow \Dcal \cup \{(\h_t, \y_t, r_t)\}$ where $\h_t$ is the discretized $h_t$ at resolution $r_t$.
		\STATE Re-train the  probabilistic multi-resolution FNO on $\Dcal$
		\ENDFOR
	\end{algorithmic}
	%\vspace{-0.05in}
\end{algorithm}
%\vspace{-0.4in}
\end{figure}

%% file: related.tex
%\section{Related Work}
\section{RELATED WORK}
Operator learning is a type  of surrogate modeling aimed at mapping an input function to an output function. 
Many prior works in surrogate modeling instead map a limited set of system parameters, \eg PDE parameters, to the output function, \eg PDE solution functions, such as~\citep{higdon2008computer,zhe2019scalable,li2021scalable,wang2021multi,xing2021residual,xing2021deep,li2022infinite}.
A variety of operator learning methods have been developed, most of which are based on neural networks and henceforth called neural operators.  For example, along with FNO, a simple low-rank neural operator (LNO)~\citep{li2020fourier} was proposed to employ a low-rank decomposition of the operator's kernel.
 \citet{li2020neural} proposed  GNO that uses Nystrom approximation and graph neural networks to approximate the function convolution. In \citep{li2020multipole}, a multipole graph neural operator (MGNO) is developed, which uses a multi-scale kernel decomposition to achieve linear complexity in computing the convolution.  \citet{gupta2021multiwavelet} developed a multiwavelet-based operator learning model that represents the operator's kernel with fine-grained wavelets.  Another popular approach is the Deep Operator Net (DeepONet)~\citep{lu2021learning}, which combines a branch net over the input functions and a trunk net over the sampling locations to predict the target function values.  A more stable and efficient version, POD-DeepONet was proposed in~\citep{lu2022comprehensive}, which replaces the trunk net with the POD (or PCA) bases computed from the training data.  A survey of neural operators is given in \citep{kovachki2023neural}. Recent work has also developed kernel operator learning approaches~\citep{long2022kernel,batlle2023kernel}.

Active learning is a classical machine learning topic. The recent research focuses on the active learning of deep neural networks. For example, in~\citep{gal2017deep}, Monte-Carlo (MC) Dropout~\citep{gal2016dropout} was used to generate the posterior samples and compute the acquisition function.\cmt{Based on the same framework, \citet{kirsch2019batchbald} developed a batch active learning method.}   \citep{geifman2017deep,sener2018active} used core-set search to query diverse and representative examples, which are shown to be particularly effective for convolution neural nets. Other examples include \citep{gissin2019discriminative,ducoffe2018adversarial} for adversarial active learning, \citep{ash2019deep} using the gradient magnitude to represent the uncertainty and to query new examples, \etc  Recently, \citep{li2022deep} proposed the first multi-fidelity active learning approach, which dynamically queries multi-fidelity simulation examples to train a surrogate model that predicts  PDE solutions from PDE parameters. \citep{li2022batch} further developed a batch multi-fidelity active learning algorithm with budget constraints.  The key difference is that these works aim to learn a mapping from the PDE parameters (low-dimensional input) to the solution (high-dimensional output), and they employ an auto-regressive architecture to combine examples of multiple fidelities. The stochastic variational inference used in the method, however, is inferior in  posterior approximation and uncertainty quantification for operator learning. We therefore develop another  posterior inference approach based on ensemble learning, which turns out to be much more effective. We accordingly develop an efficient method for utility function computation. In addition, we discovered the over-penalization problem during the active learning, which was never discovered in these previous works. We proposed a novel and flexible cost annealing framework to overcome the problem. The recent work~\citep{pickering2022discovering} proposed an active learning approach for DeepONet. The goal is to query examples that can facilitate the discovery of rare events hidden in physical systems. The work does not consider multi-resolution examples and their varying costs. The goal, model estimation, acquisition function design and computation are all different from our work.

%% file: exp-zhe.tex
\section{EXPERIMENT}

\subsection{Prediction Accuracy on Fixed Training Data}\label{sect:expr:fixed}
We first examined if our multi-resolution FNO can achieve good prediction accuracy and uncertainty calibration. To this end, we tested with two benchmark operator learning tasks, one is based on a Burgers' equation and the other a Darcy flow equation. For \textit{Burgers}, we aim to learn a mapping from the initial condition to the solution at time $t=1$, while for \textit{Darcy}, the goal is to learn a mapping from the coefficient function to the solution. We considered two resolutions for each task. The details are provided in Section \ref{sect:app:detail} in Appendix. 
\begin{table}[]
	\centering
	\small
	\begin{subtable}{0.5\textwidth}
		\small
		\centering
	\begin{tabular}{cccc}
		\hline
		{Method}      &  & {Burgers}             & {Darcy}               \\ \hline
		FNO                  &  & \textbf{0.0575 $\pm$ 0.0031} & 0.0891 $\pm$ 0.0078          \\
		FNO-Dropout          &  & 0.0791 $\pm$ 0.0035          & 0.1038 $\pm$ 0.0056          \\
		FNO-SGLD             &  & 0.0804 $\pm$ 0.0049          & 0.0933 $\pm$ 0.0074          \\
		FNO-SVI              &  & 0.1182 $\pm$ 0.0056          & 0.0946 $\pm$ 0.0041          \\
		\ours &  & \textbf{0.0586 $\pm$ 0.0042} & \textbf{0.0876 $\pm$ 0.0059} \\ \hline
	\end{tabular}
	\caption{\small Relative $L_2$ error}\label{tb:nac-l2}
	\end{subtable}
	\begin{subtable}{0.5\textwidth}
		\small
		\centering
		\begin{tabular}{c|ccc}
			\hline
			{Method}      & {} & {Burgers}          & {Darcy}               \\ \hline
			FNO                  &           & NA                        & NA                           \\
			FNO-Dropout          &           & 176.84 $\pm$ 16.11        & 4447.92 $\pm$63.08           \\
			FNO-SGLD             &           & 223.48 $\pm$ 15.74        & 3683.45 $\pm$ 83.49          \\
			FNO-SVI              &           & 391.61 $\pm$ 10.59        & 4027.71 $\pm$ 73.96          \\
			\ours &           & \textbf{44.57 $\pm$ 3.62} & \textbf{1167.73 $\pm$ 29.13} \\ \hline
		\end{tabular}
	\caption{\small Negative Log-Likelihood (NLL)}\label{tb:nac-nll}
	\end{subtable}
	\begin{subtable}{0.5\textwidth}
	\small
	\centering
	\begin{tabular}{ccc}
		\hline
		{Method}      & {Burgers}             & {Darcy}               \\ \hline
		FNO                   & 296.31\% & 218.54\%         \\
		FNO-Dropout          &    331.64\%      & 233.04\%          \\
		FNO-SGLD               & 200.62\%         &183.60\%         \\
		FNO-SVI                & 111.14\%         & 188.16\%          \\
		\ours &  286.86\% & 201.71\% \\ \hline
	\end{tabular}
	\caption{\small Relative $L_2$ error increase}\label{tb:nac-increase}
	\end{subtable}
	\caption{\small Prediction accuracy using 200 high-resolution and 200 low-resolution examples (a, b), and compared to (a) the error increase of using 400 low-resolution examples (c). The results were averaged from five runs. } \label{tb:surrogate_rmse}
	\end{table}

\begin{figure*}[t]
	\centering
	\setlength\tabcolsep{0pt}
	\begin{tabular}[c]{ccc}
		\setcounter{subfigure}{0}
		\begin{subfigure}[t]{0.33\textwidth}
			\centering
			\includegraphics[width=\textwidth]{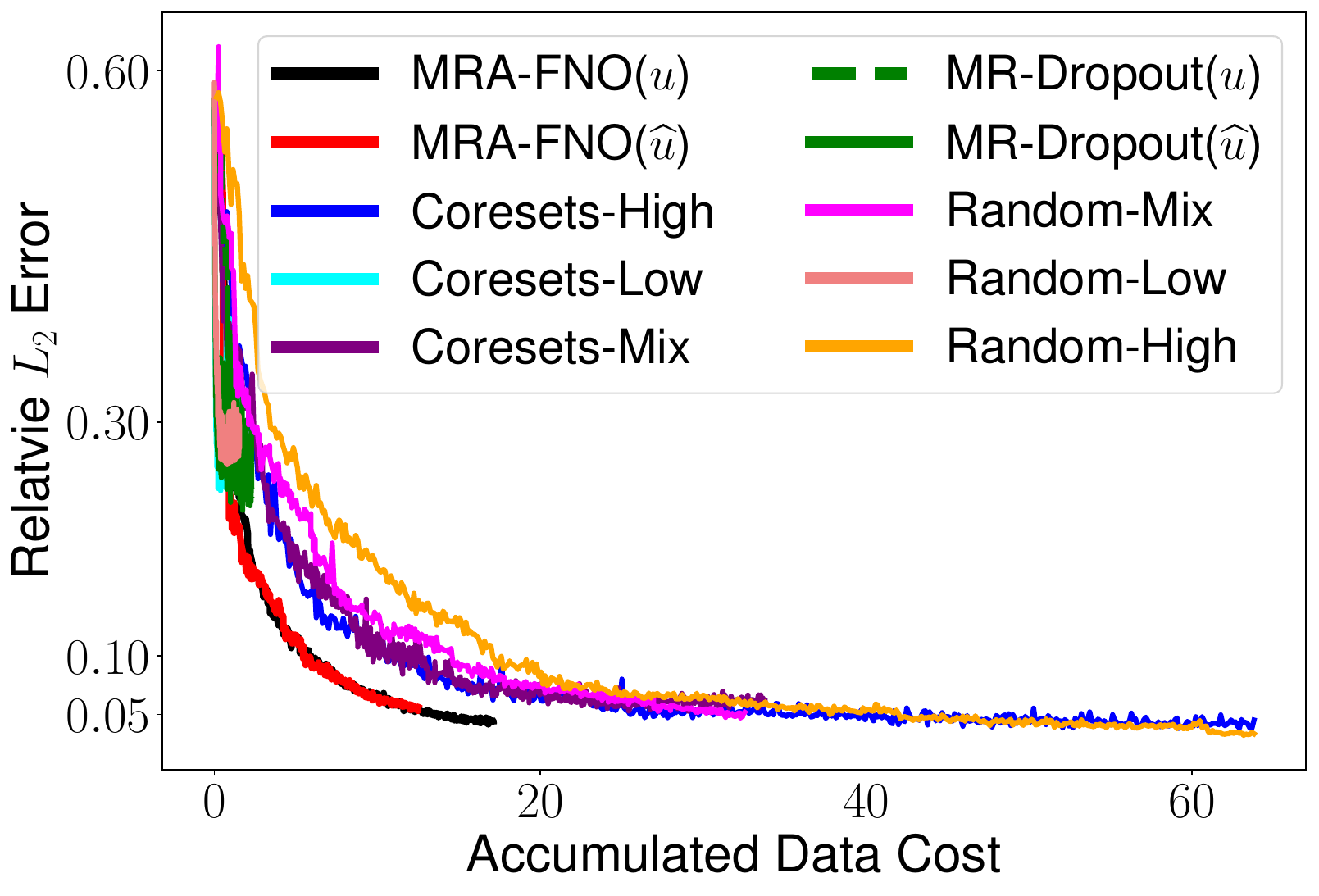}
			\caption{\small \textit{Burgers}}
		\end{subfigure}
		&
		\begin{subfigure}[t]{0.33\textwidth}
			\centering
			\includegraphics[width=\textwidth]{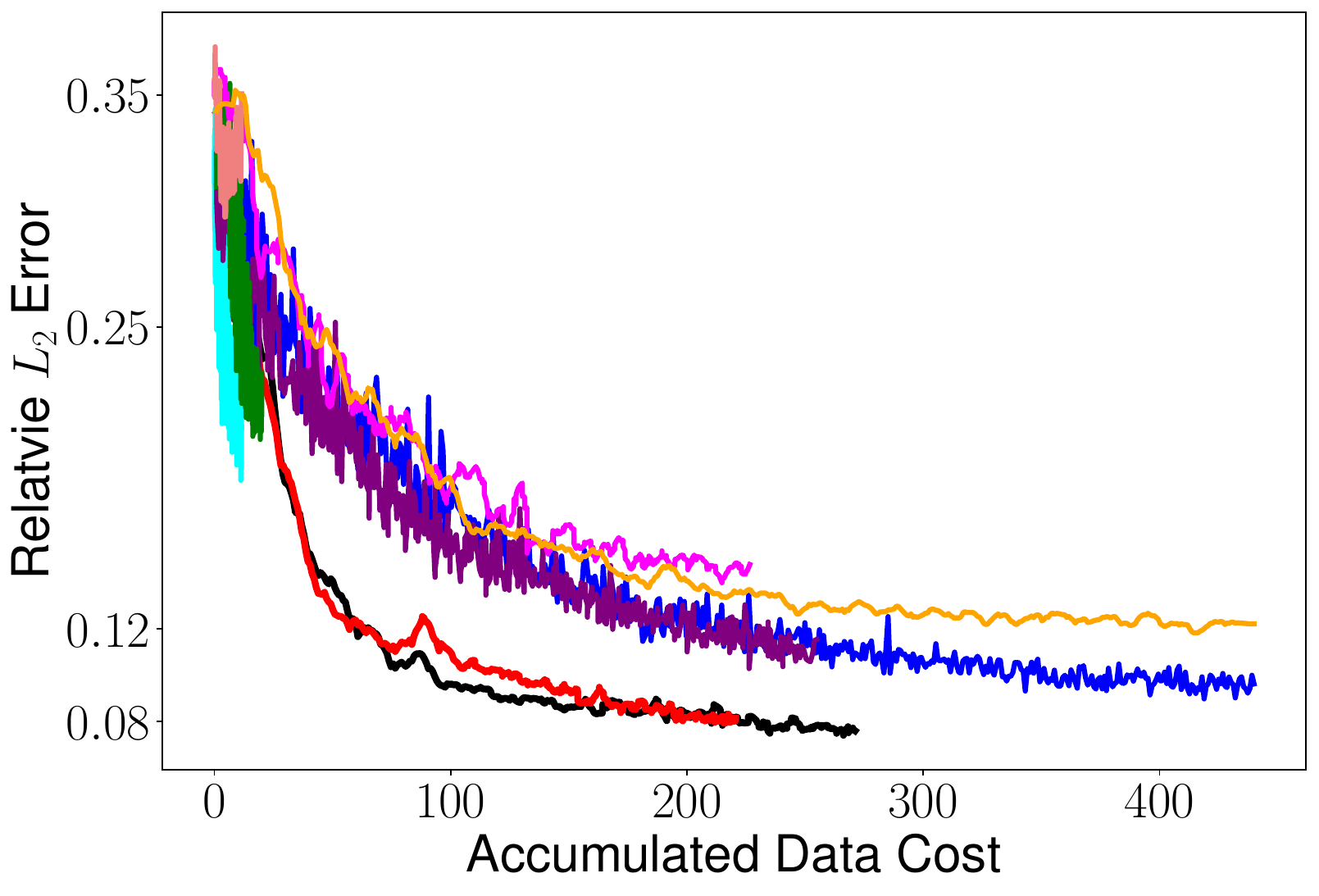}
			\caption{\small \textit{Darcy}}\label{fig:accuracy-exp-darcy}
		\end{subfigure}
		&
		\begin{subfigure}[t]{0.33\textwidth}
			\centering
			\includegraphics[width=\textwidth]{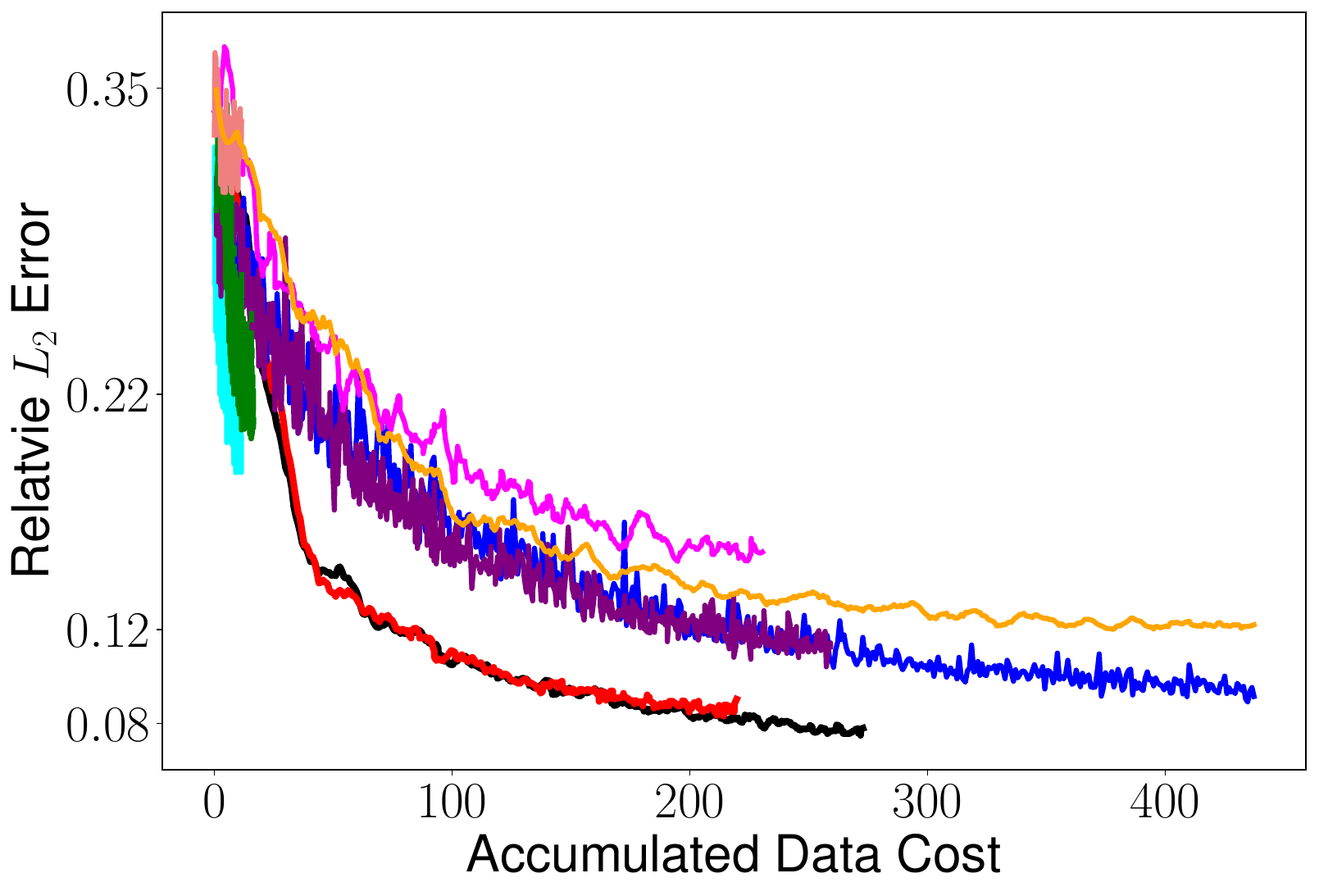}
			\caption{\small \textit{Darcy3}}\label{fig:accuracy-exp-darcy3}
		\end{subfigure}
	\end{tabular}
	\caption{\small Relative $L_2$ error \textit{vs.} accumulated data cost. Each method ran 500 active learning steps.  Note that different methods can end up with different total data cost (after running the same number of steps).}  \label{fig:accuracy-exp}
\end{figure*}
\begin{figure*}[t]
	\centering
	\setlength\tabcolsep{0pt}
	\begin{tabular}[c]{cccc}
		\begin{subfigure}[t]{0.24\textwidth}
			\centering
			\includegraphics[width=\textwidth]{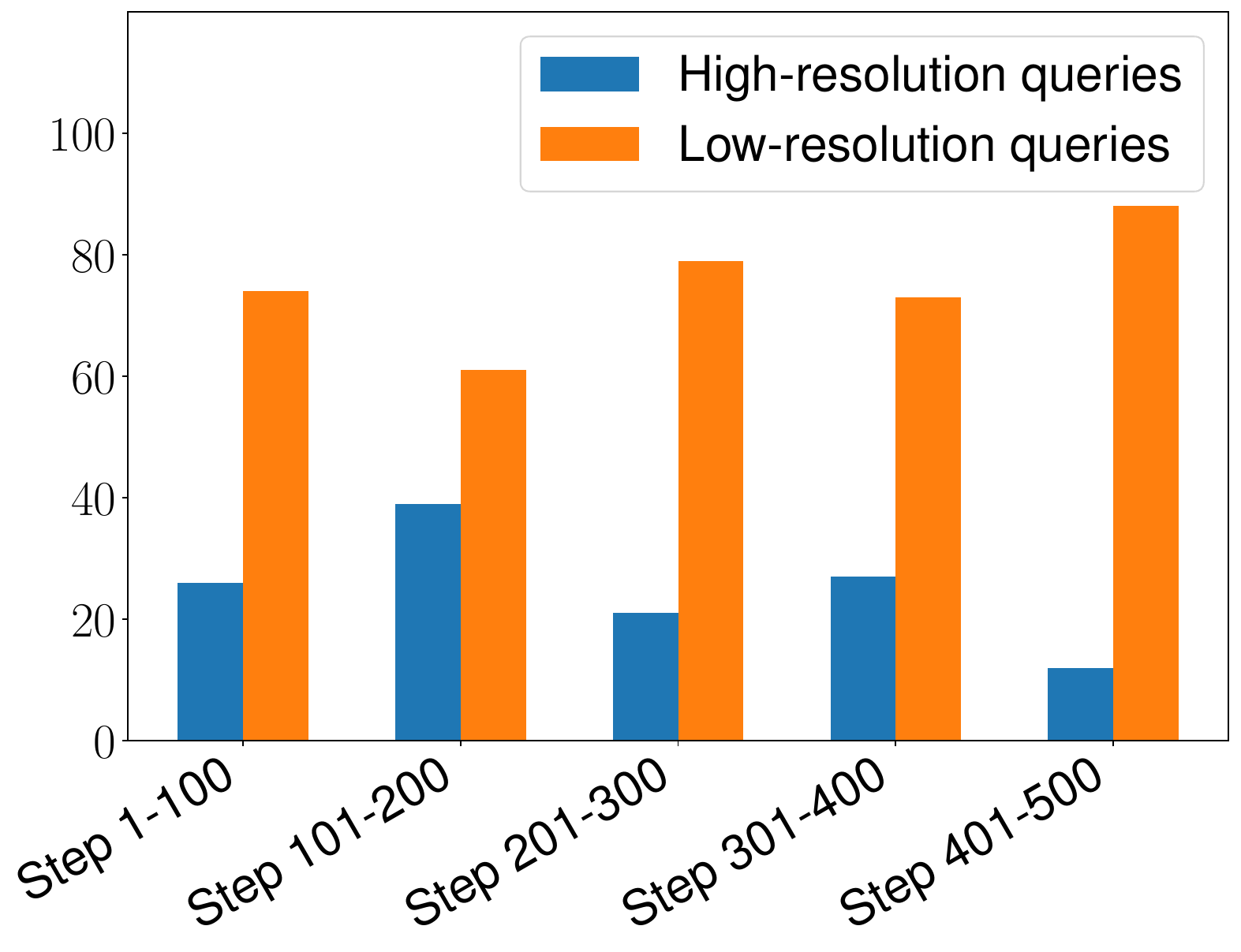}
			\caption{\small \textit{Burgers}}
		\end{subfigure}
		&
		\begin{subfigure}[t]{0.24\textwidth}
			\centering
			\includegraphics[width=\textwidth]{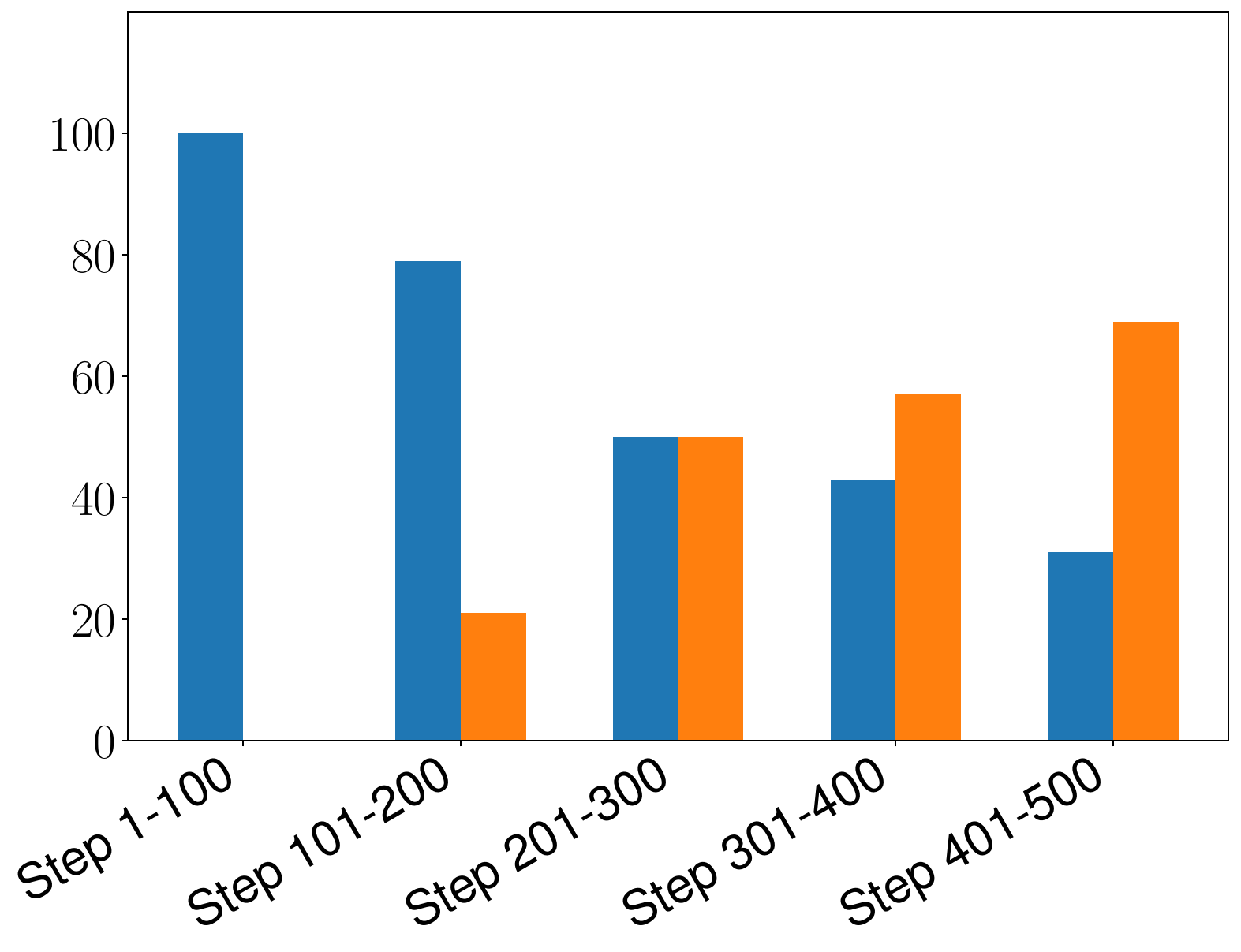}
			\caption{\small \textit{Darcy}}
		\end{subfigure}
		&
		\begin{subfigure}[t]{0.24\textwidth}
			\centering
			\includegraphics[width=\textwidth]{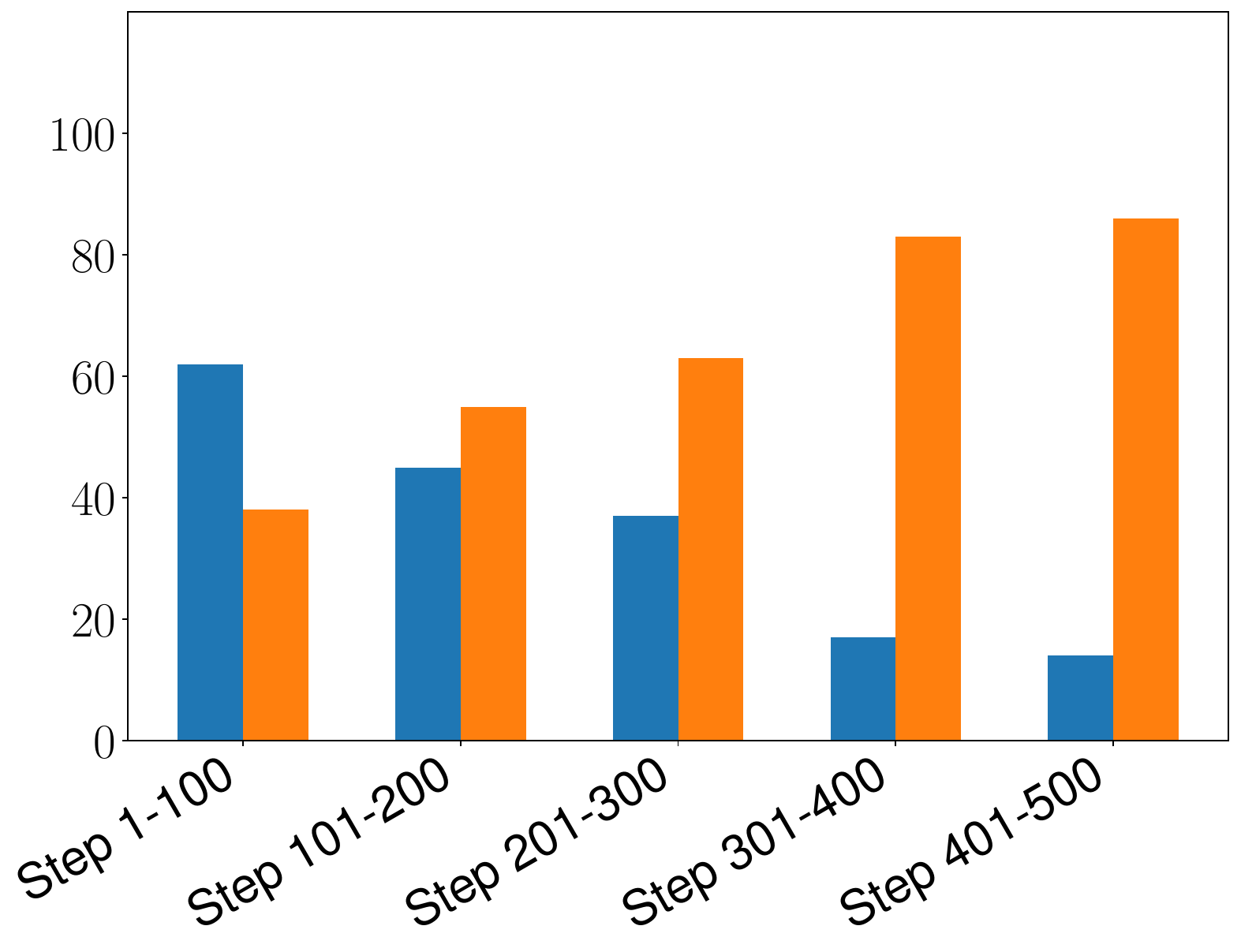}
			\caption{\small \textit{Diffusion}}
		\end{subfigure}
		&
		\begin{subfigure}[t]{0.23\textwidth}
			\centering
			\includegraphics[width=\textwidth]{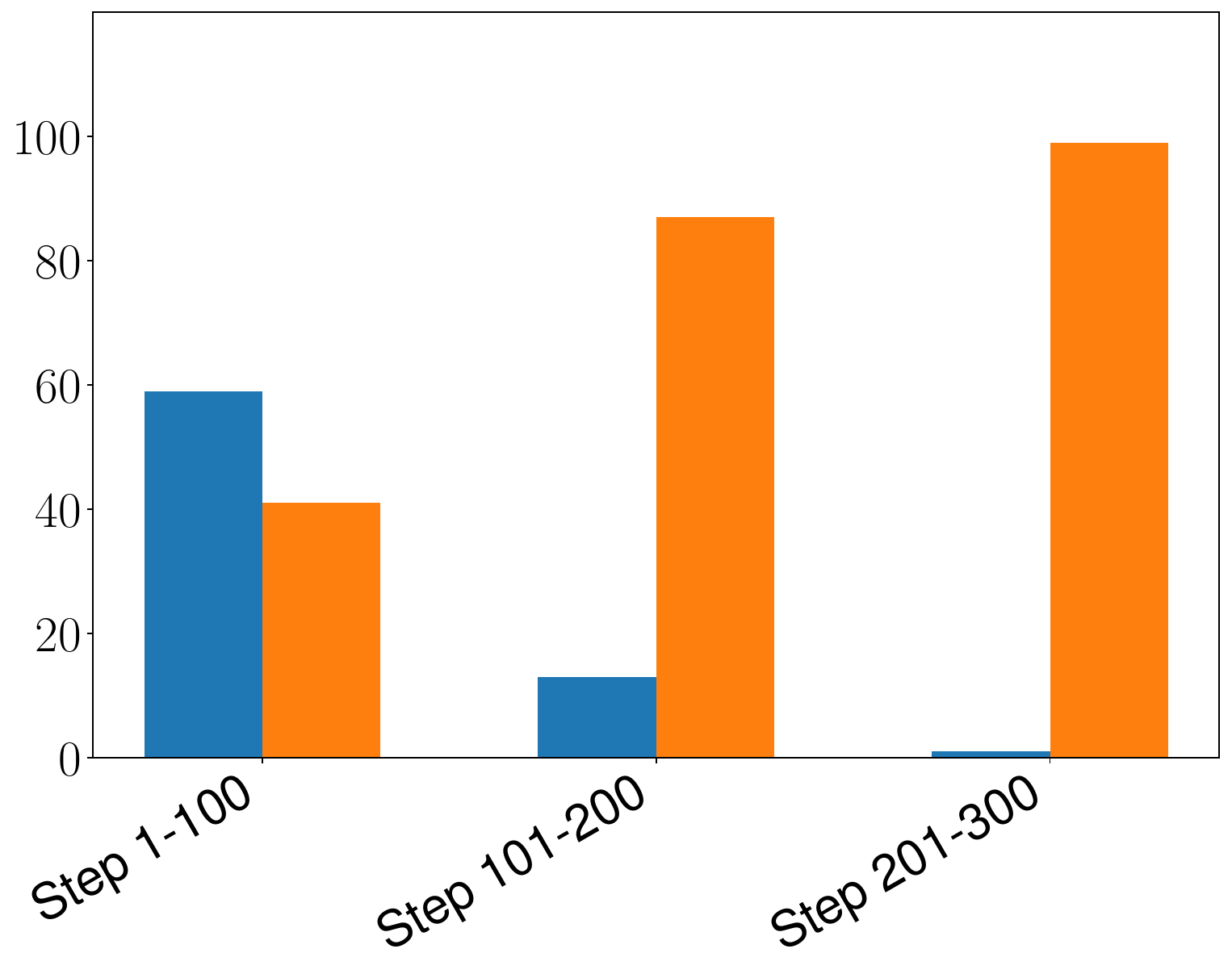}
			\caption{\small \textit{Navier Stoke (NS)} }
		\end{subfigure}
	\end{tabular}
	%	\vspace{-0.1in}
	\caption{\small Number of resolutions queried by \ours at different stages during active learning.} \label{fig:query-count}
\end{figure*}
\begin{figure}%{r}{0.66\textwidth}
	\centering
	\setlength\tabcolsep{0pt}
	\begin{tabular}[c]{c}
		\setcounter{subfigure}{0}
		\begin{subfigure}[t]{0.33\textwidth}
			\centering
			\includegraphics[width=\textwidth]{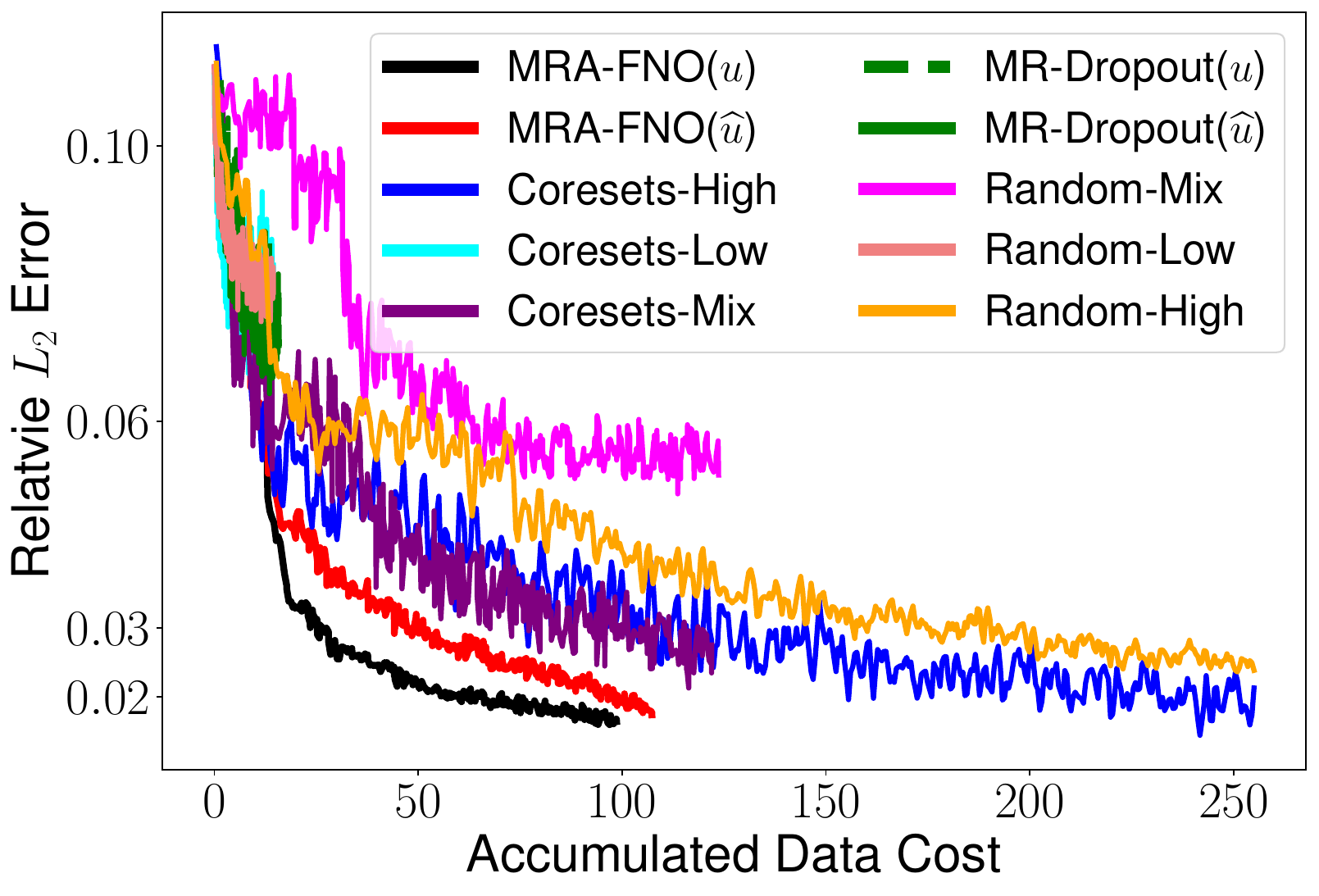}
			\caption{\small \textit{Diffusion}}
		\end{subfigure}
		\\
		\begin{subfigure}[t]{0.33\textwidth}
			\centering
			\includegraphics[width=\textwidth]{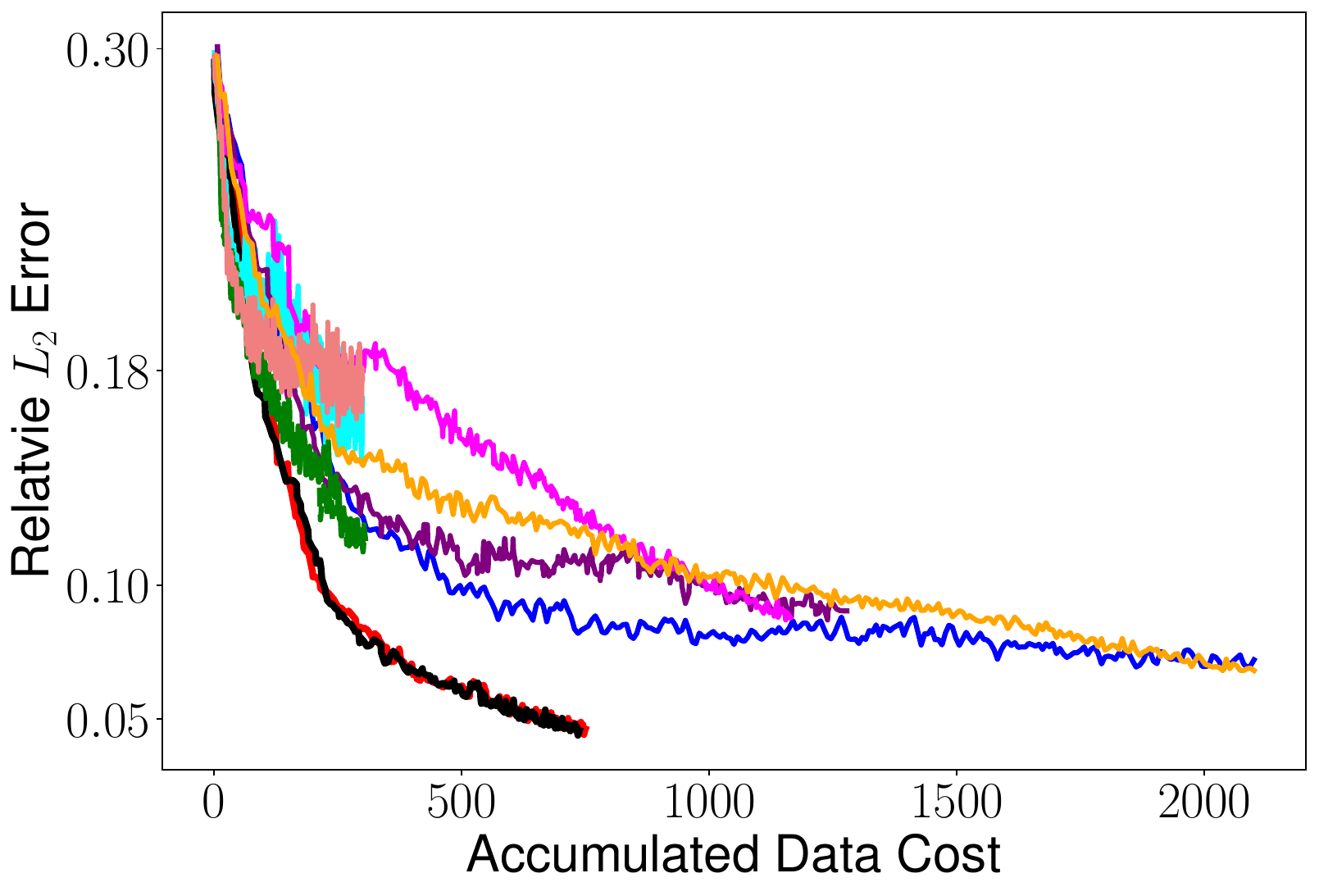}
			\caption{\small \textit{Navier Stoke (NS)}}
		\end{subfigure}
	\end{tabular}
	\caption{\small Relative $L_2$ error \textit{vs.} accumulated data cost. } \label{fig:acc-exp2}
\end{figure}
 We randomly generated 200 examples for each resolution to obtain a training set. We randomly generated another 200 examples at the highest resolution as the test set.  We compared with the standard FNO (point estimation),  FNO trained via MC Dropout (FNO-Dropout)~\citep{gal2016dropout}, stochastic gradient Langevin dynamics (FNO-SGLD)~\citep{welling2011bayesian}, and stochastic variational inference (FNO-SVI)~\citep{kingma2013auto}.  For all the methods, we set the mini-batch size to $20$,  the learning rate to $10^{-3}$, and use ADAM optimization and Cosine Annealing schedule. We used the FNO implementation from the original authors (\url{https://github.com/neuraloperator/neuraloperator}).   
 We tuned the dropout rate from \{0.1, 0.2, 0.3, 0.4, 0.5\}. For SGLD and SVI, we assigned a standard Gaussian prior over the model parameters. For SVI, we employed a fully factorized Gaussian posterior approximation. 
We repeated the training and test procedure for five times, and examined the average relative $L_2$ error, the average negative log likelihood (NLL), and their standard deviation on the test datasets. The results  are reported in Table \ref{tb:nac-l2} and \ref{tb:nac-nll}.  To confirm the value of high-resolution examples, we also ran all the methods using 400 low resolution examples, and tested if the error will increase. The results are summarized in  Table \ref{tb:nac-increase}.

From Table \ref{tb:nac-l2} and \ref{tb:nac-nll}, one can see that the our model (\ours) achieves the relative $L_2$ error  significantly smaller than the competing methods in all the cases,  except that in Burger's equation, the $L_2$ error of \ours is slightly worse than the standard FNO. More important, \ours consistently outperforms all the probabilistic versions of FNO by a large margin in  test log likelihood.  Hence, not only does our model give superior prediction accuracy, our ensemble posterior inference also enables much better uncertainty quantification. 

Table \ref{tb:nac-increase} illustrates that, without high-resolution examples, the relative $L_2$ error of all the methods increases dramatically, even with the same training size. This underscores the importance of high-resolution examples for achieving good accuracy, despite their higher cost of collection.

\subsection{Active Learning Performance}
Next, we evaluated the active learning performance of \ours. In addition to the tasks in Section  \ref{sect:expr:fixed}, we considered two more PDEs, one is a nonlinear diffusion equation, and the other is a 2D Navier-Stokes (NS) equation used in~\citep{li2020fourier}. For each task, we considered two resolutions. We leave the details in Section \ref{sect:app:detail} of Appendix.  In addition, we tested active learning on the same \textit{Darcy} problem as in Section \ref{sect:expr:fixed} with three resolutions. We summarize the data acquiring cost at different resolutions in Table \ref{tb:task-summary}.  As we can see, the cost discrepancy is large among different resolutions. 
\begin{table}[]
%\begin{wraptable}{r}{0.57\textwidth}
	\small
	\centering
	\begin{tabular}{c|cc}
		\hline
		{Task}      & Resolution & Cost Ratio\cmt{ $\lambda_1 : \lambda_2 (: \lambda_3)$} \\ \hline
		\textit{Burgers}            & 33, 129 & $1 : 41.2$                       \\
		\textit{Darcy}             & $32\times 32$, $128 \times 128$ & $1 : 38.3$                       \\
		\textit{Darcy3}             & $32\times 32$, $64\times 64$, $128 \times 128$ & $1 : 21.3 : 38.3$                \\
		\textit{Diffusion} & $32\times 32$,  $128 \times 128$& $1 :  17.6$                 \\
		\textit{NS}       & $16 \times 16$, $64 \times 64$ & $1 : 7$                          \\ \hline
	\end{tabular}
\caption{\small Resolution and cost ratio for each active learning task. The cost is measured by the average running time for solving the PDEs (100 runs) at the corresponding resolution. }\label{tb:task-summary}
%\end{wraptable}
\end{table}
We compared with the following active learning methods for FNO: (1) Random-Low/High, randomly selecting an input function from the candidate pool, and querying the example at the lowest/highest resolution. (2) Random-Mix, randomly selecting both the input and resolution. (3)  Coreset-Low/High, using the coreset active learning strategy~\citep{sener2018active} to select the input function that maximizes the minimum distance to the existed examples, according to the output of the last Fourier layer as the representation. We fixed the resolution to be the lowest or the highest one. (4) Coreset-Mix, the same coreset active learning strategy as in (3), except that we allow querying at different resolutions.  We interpolate the representation to the highest resolution to compute the distance. (5) MR-Dropout: we used MC dropout to perform posterior inference for FNO, and then used the same acquisition function(s), computation method, and annealing framework as in our approach to identify the input function and resolution. (6) MR-PredVar: we averaged the predictive variance of each output function values as the utility function, and the remaining is the same as our approach.  We ran all our experiments on Lambda Cloud instances equipped with A100 GPUs. 

\begin{figure*}[t]
	\centering
	\setlength\tabcolsep{0pt}
	\begin{tabular}[c]{ccc}
		\setcounter{subfigure}{0}
		\begin{subfigure}[t]{0.31\textwidth}
			\centering
			\includegraphics[width=\textwidth]{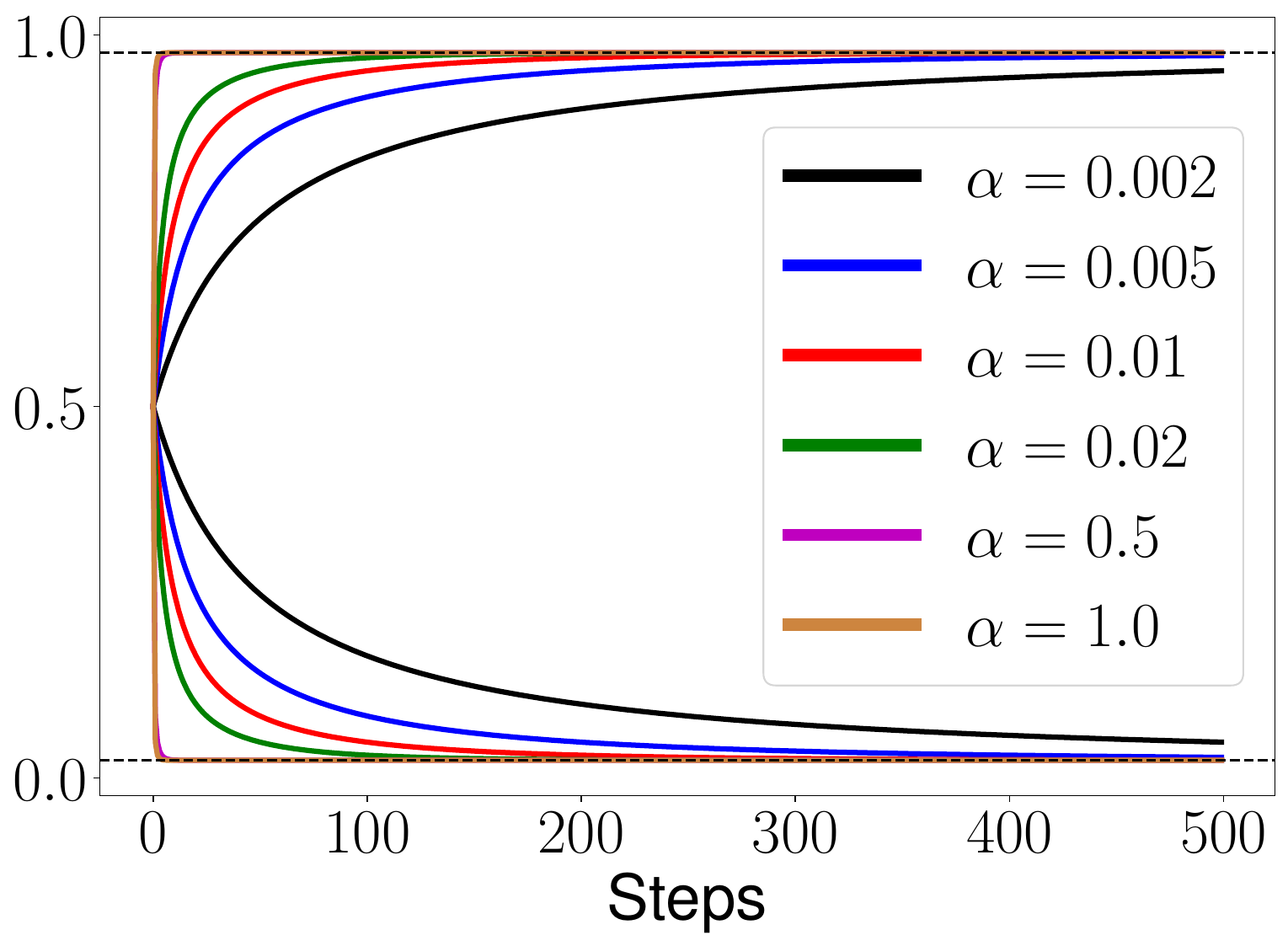}
			\caption{\small Cost schedule (Exp)} \label{fig:schedule-exp}
		\end{subfigure}
		&
		\begin{subfigure}[t]{0.33\textwidth}
			\centering
			\includegraphics[width=\textwidth]{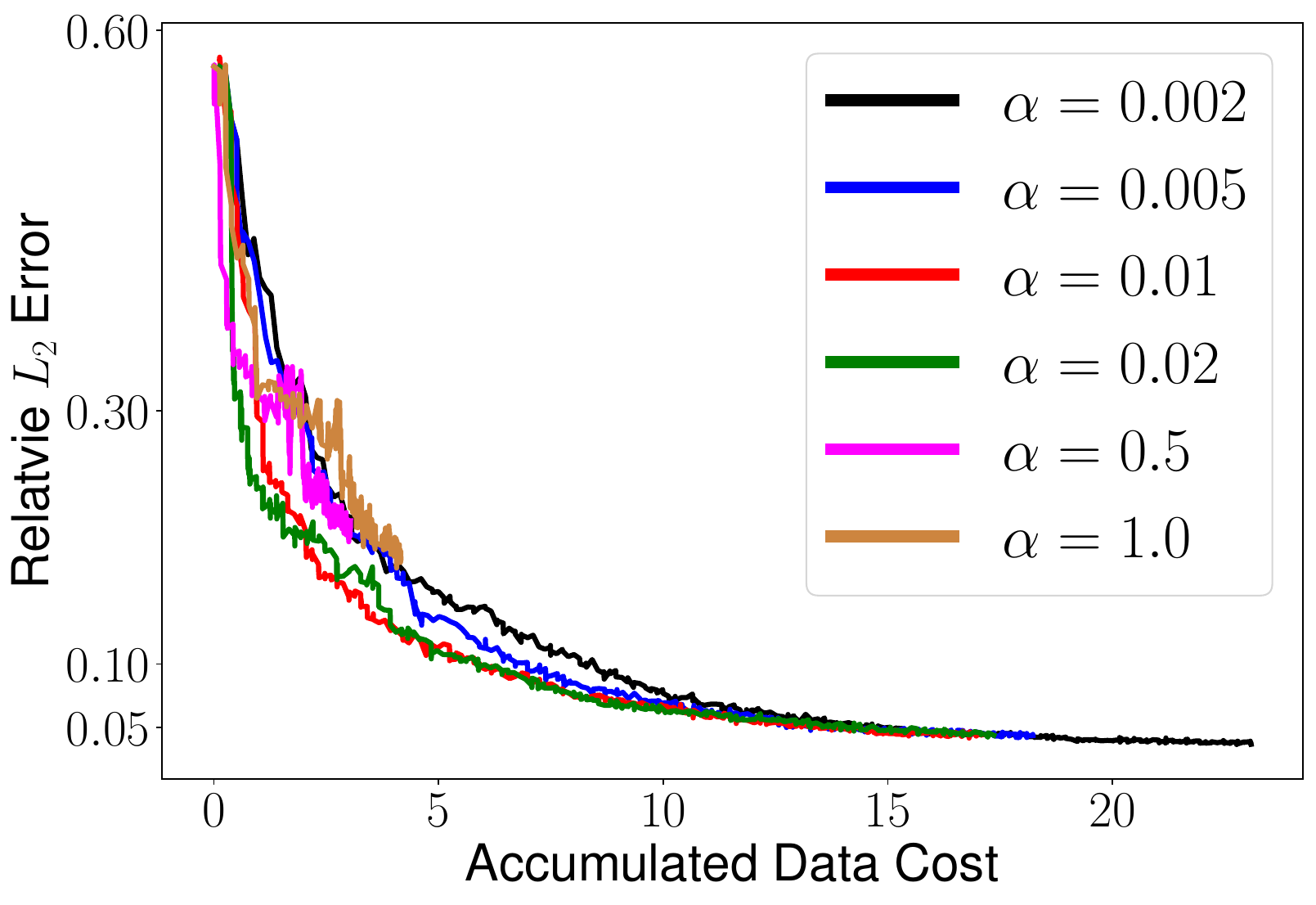}
			\caption{\small \ours($u$)}\label{fig:u-sche}
		\end{subfigure}
		&
		\begin{subfigure}[t]{0.33\textwidth}
			\centering
			\includegraphics[width=\textwidth]{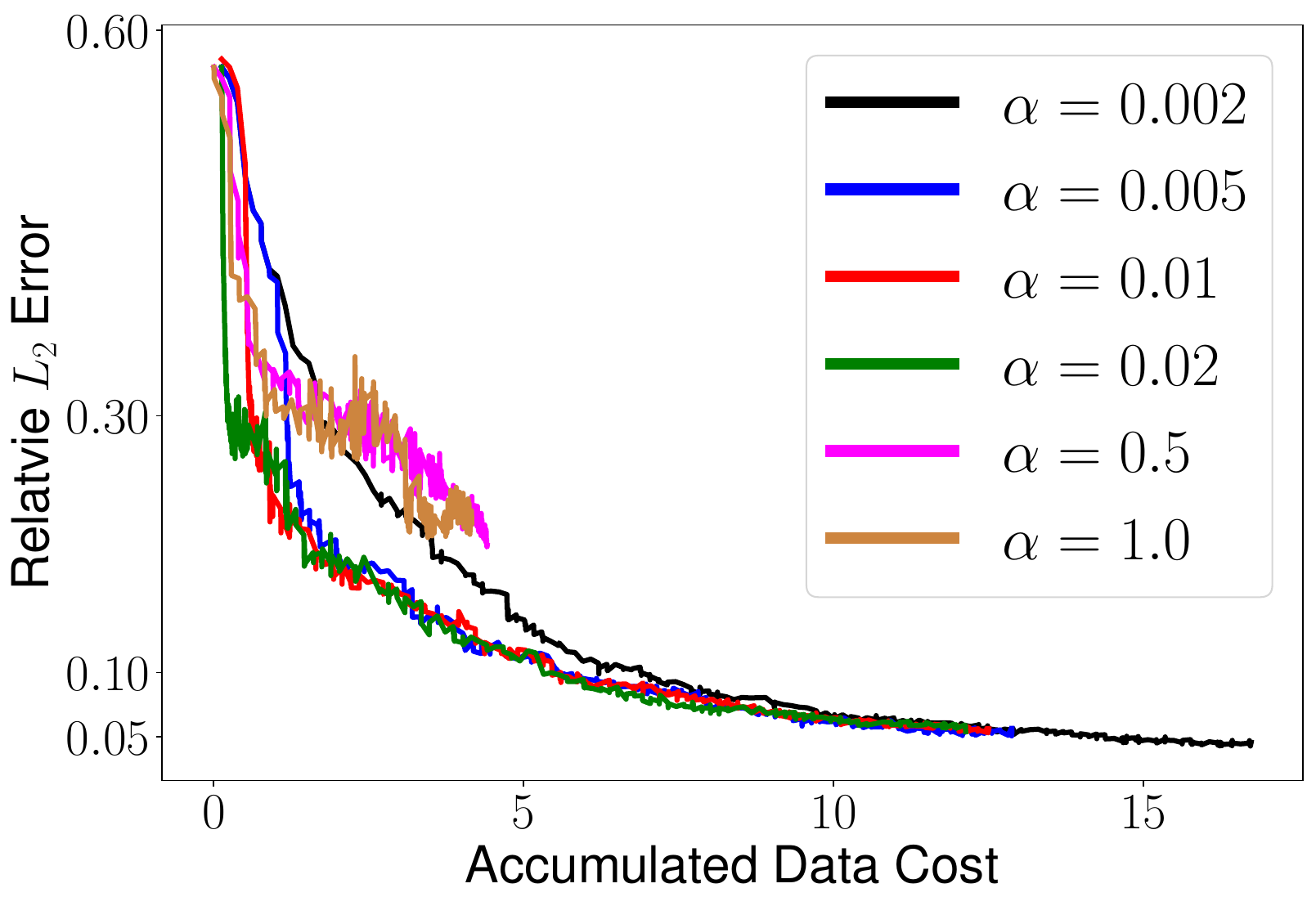}
			\caption{\small \ours($\widehat{u}$)}\label{fig:uhat-sche}
		\end{subfigure}
	\end{tabular}
	\caption{\small The influence of the cost schedule on active learning. We report the result with the exponential decay; see \eqref{eq:decay-func}.  The larger $\alpha$, the faster the schedule converges to the true cost. } 
\end{figure*}
\begin{figure*}
	\centering
	\setlength\tabcolsep{0pt}
	\includegraphics[width=0.7\textwidth]{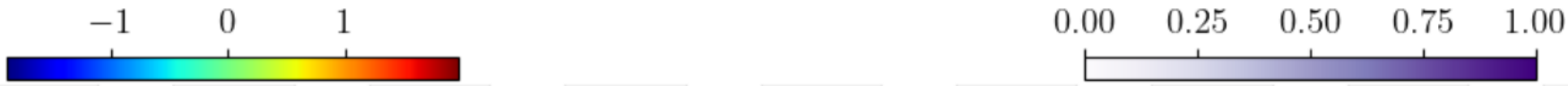}
	\begin{tabular}[c]{c}
		\setcounter{subfigure}{0}
		\begin{subfigure}[t]{0.8\textwidth}
			\centering
			\includegraphics[width=\textwidth]{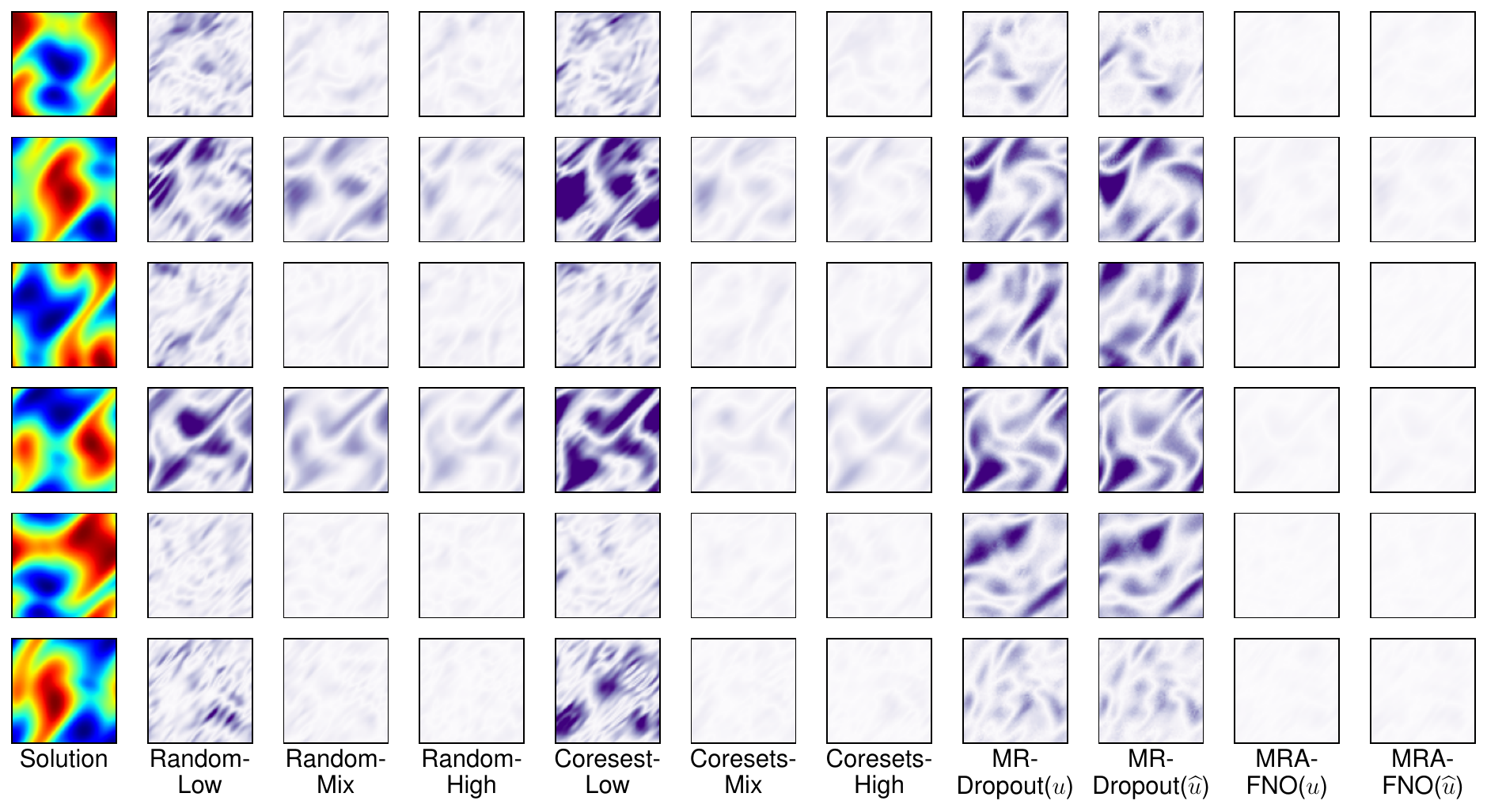}
		\end{subfigure}
	\end{tabular}
	\caption{\small Point-wise error on \textit{NS}.} \label{fig:point-wise-ns}
\end{figure*}
For every active learning experiment, we randomly generated 10 examples for each resolution to obtain an initial dataset. We randomly generated 990 input functions at the highest resolution, which we used as the candidate pool for active learning. If one example is queried at a lower resolution, the  input function is downsampled accordingly at which to run the simulation. We randomly generated another 200 examples at the highest resolution for testing. 
We then ran active learning with each method. For our method and MR-Dropout, we tested two annealing schedules, one is based on the exponential decay and the other sigmoid decay; see \eqref{eq:decay-func} . We tuned the decaying rate $\alpha$ from $\{0.002, 0.005, 0.01, 0.02, 0.5, 1.0\}$. We ran $500$ active learning steps (queries) for all the experiments except for the \textit{NS} problem, we ran $300$ steps. We examined the relative $L_2$ error of each method \textit{vs.} the accumulated data cost. To avoid cluttered figures, we show the result of our method with the exponential-decay-based schedule in Fig. \ref{fig:accuracy-exp} and \ref{fig:acc-exp2}, and the result of using the sigmoid decay and MR-PredVar in Fig. \ref{fig:acc-more} in Appendix.

\textbf{Prediction Accuracy.} As we can see, at the beginning, the performance of each method is identical or very close. As the active learning progresses, \ours improves rapidly and constantly. It soon achieves a superior prediction accuracy to all the competing methods, and consistently outperforms them during the remaining course of the active learning. Accordingly, \ours can reach the smallest prediction error under the same data cost, or use the least data cost to achieve the same performance. We empirically observed that  using the utility function \eqref{eq:u1} or \eqref{eq:u2}, denoted by \ours ($u$) and \ours ($\widehat{u}$), respectively,  result in close performance, except that on the diffusion problem, \ours ($u$) appears to be better. This might be because the Monte-Carlo approximation in \eqref{eq:u2} (we set $A = 5$) still has a significant gap from the true expectation.  It is worth noting that both Random-Low and Coreset-Low were quickly trapped  at large prediction errors. It therefore shows only using low-resolution examples, the predictive performance will soon meet a bottleneck and can hardly improve, though the data cost grows very slowly. On the other hand, Random-High and Coreset-High  enables steady improvement because they only query high-resolution examples at each step. However, the data cost accumulation is much greater, \eg Fig. \ref{fig:accuracy-exp-darcy} and \ref{fig:accuracy-exp-darcy3}. In addition, the performance of MR-Dropout tends to stuck at large prediction errors early, especially in \textit{Burgers}, \textit{Darcy} and \textit{Darcy3}. We observed that  MR-Dropout mainly selected low-resolution examples. This might be because the uncertainty quantification by dropout is not reliable for FNO, and even using our annealing framework cannot correct its bias. From Fig. \ref{fig:acc-more} of Appendix, we can see that the performance of \ours with the sigmoid-based cost schedule is close to that with exp-based schedule (see \eqref{eq:decay-func}), except in \textit{Darcy3}, the exp-based schedule shows a slight yet consistent advantage. Interestingly, MR-PredVar outperforms the other competing methods in all the cases, confirming the importance of effective uncertainty quantification in utility evaluation (it also uses our ensemble posterior inference). While MR-PredVar achives close performance to our method in 
\textit{Burgers}, in all the other cases, MR-PredVar is apparently worse. This might be because MR-PredVar ignores the (strong) correlation between the output function values, and hence the quality of utility evaluation is worse.  

In addition, we examined the count of queried resolutions at different stages by \ours ($u$) with the exp-based schedule. As depicted in Fig. \ref{fig:query-count}, during the first 100 or 200 steps, \ours queried the majority of high-resolution examples. Subsequently, \ours increasingly queried low-resolution examples, while the test performance continued to improve, consistently outperforming other methods by a large margin. These findings validate the effectiveness of our cost annealing schedule. Overall, these results demonstrate the advantage of our multi-resolution active learning approach.

\textbf{Influence of Cost Schedule.} Next, we investigated how the cost annealing schedule influences the active learning. To this end, we used the exponential decay function in our schedule, and varied the decaying rate $\alpha \in \{0.002, 0.005, 0.01, 0.02, 0.5, 1.0\}$. We show the cost schedule for different choices of $\alpha$ in Fig. \ref{fig:schedule-exp}.  We then run \ours on \textit{Burgers} with 500 steps. The $L_2$ relative error \textit{vs.} the accumulated data cost is reported in Fig. \ref{fig:u-sche} and \ref{fig:uhat-sche}. We can see that when $\alpha$ is too small, \eg $\alpha = 0.002$, though the active learning ensures steady improvement of the prediction accuracy, the data cost is suboptimal. To obtain the same performance, a too small $\alpha$ consumes a much bigger data cost, or under the same cost, it gives worse performance. The reason is that the convergence of the cost annealing is too slow; see Fig. \ref{fig:schedule-exp}. Even when the mutual information has become sufficiently discriminative, the cost assignments for different resolutions are still not far, which actually over-penalize low-resolution examples and lead to a selection bias toward high-resolution examples. Another extreme is to use a too big $\alpha$, \eg $\alpha = 0.5$ and $\alpha=1.0$. In such case, the schedule will converge to the true cost very fast, even at the early  stage when data is few. Accordingly, the high-resolution examples are soon over-penalized, making the learning stuck at low-resolution queries. The prediction accuracy is fluctuating yet hard to increase substantially. On the contrary, an appropriate decay rate in between, \eg $\alpha = 0.01$ and $\alpha=0.02$, can sidestep these problems, and lead to superior performance in both cost saving and prediction accuracy. 

\textbf{Point-wise Error.} Finally, we investigate the local errors of the prediction. We randomly selected six test cases for \textit{NS} and \textit{Diffusion}. We examined the post-wise error of each method after active learning. \ours used the exp-based schedule. We show the results in Fig. \ref{fig:point-wise-ns} and Appendix Fig. \ref{fig:point-wise-diffusion}. We can see that the point-wise error of \ours is quite uniform across the domain and is close to zero (white). By contrast, the other methods exhibit  large errors in many local regions.  Together these results have shown that \ours not only gives a superior global accuracy, but locally better recovers individual output function  values.

%% file: conclusion.tex
%\vspace{-0.1in}
%\section{Conclusion}
\section{CONCLUSION}
%\vspace{-0.1in}
We have presented \ours, a multi-resolution active learning method for Fourier neural operators. On several benchmark operator learning tasks, \ours can save the data cost substantially while achieving superior predictive performance. Currently, the selection of the decay rate in our cost annealing framework is done by manual tuning or cross-validation. In the future, we plan to develop novel methods, such as reinforcement learning, to automatically determine the best rate.

\section*{Acknowledgments}
This work has been supported by MURI AFOSR
grant FA9550-20-1-0358, NSF CAREER Award
IIS-2046295, and NSF OAC-2311685.  We thank Andrew Stuart for valuable discussions and suggestions.

%% file: suppl-cr.tex
% Supplementary material: To improve readability, you must use a single-column format for the supplementary material.
\onecolumn
\section{APPENDIX}
\subsection{{OPERATOR LEARNING TASK DETAILS}}\label{sect:app:detail}
We tested our method with the following operator learning tasks.
\begin{itemize}
	\item \textbf{Burgers}. The first one is based on the Burger's equation, 
	\begin{align}
		u_t + u_{xx} = \nu u_{xx}, \;\; u(x, 0) = u_0(x) , \label{eq:burgers}
	\end{align}
	where $(x, t) \in [0, 1]^2$, and $u_0(x)$ is the initial condition, and $\nu=0.002$ is the viscosity. We aim to learn a mapping from the initial condition to the solution at $t=1$, namely, $u_0 \rightarrow u(x, 1)$. We considered two resolutions, which use 33 and 129 samples for discretization, respectively.  We used a parametric form of the input function, $u_0(x)= a\exp(-ax)\sin(2\pi x)\cos(b \pi x)$, and we then randomly sampled $a, b \in [1, 6]$ to obtain the instances. 
	\item \textbf{Darcy}. The second task is based on a 2D Darcy flow equation, 
	\begin{align}
		-\nabla (c(\x) \nabla u(\x)) = f(\x),  \label{eq:darcy}
	\end{align}
	where $\x \in [0, 1]^2$, $f(\x)=1$ is a constant forcing function, $c(\x)>0$ is the diffusion coefficient function,  and on the boundary, $u(\x) = 0$. We aim to learn the mapping from the coefficient function to the solution,  $c \rightarrow u$. We employed two sampling resolutions, $32 \times 32$ and $128 \times 128$. We followed~\citep{li2020fourier} to first sample a discretized function from a Gauss random field, and then threshold the values to be 4 or 12 to obtain the input function. 
	\item \textbf{Diffusion}. The third one is based on a nonlinear diffusion PDE, 
	\begin{align}
		u_t = 0.01u_{xx} + 0.01u^2 + f(x) \label{eq:diffu}
	\end{align}
	where $(x, t) \in (0, 1) \times (0, 1]$, $u(0, t) = u(1, t) = 0$, $u(x, 0) = 0$ and $f(x)$ is the forcing function. The goal is to learn the mapping from the forcing function to the solution, $f \rightarrow u$. We employed two resolutions for data acquiring, $32\times 32$ and $128 \times 128$. We drew samples of $f$ from a Gaussian process with an RBF kernel. Note that to use FNO and \ours, we replicated the spatial discretization of $f$ along the time dimension (steps).  
	\item \textbf{Navier Stoke (NS)}. The last task is based the a 2D Navier-Stokes (NS) equation used in~\citep{li2020fourier}. The solution $u(\x, t)$ is the vorticity of a viscous, incompressible fluid, where  $\x \in [0, 1]^2$ and $t \in [0, 50]$. We set the viscosity to $10^{-3}$. Following~\citep{li2020fourier}, we considered 40 steps in the time domain. We used the the solution at the first 20 time steps to predict the solution at the next 20 steps. For data collection, we used two resolutions $16 \times 16$ and $64 \times 64$ in the spatial domain.  We sampled the input functions from a Gaussian random field. 
\end{itemize}
\begin{figure*}[h!]
	\centering
	\setlength\tabcolsep{0pt}
	\begin{tabular}[c]{c}
		\setcounter{subfigure}{0}
		\begin{subfigure}[t]{1.0\textwidth}
			\centering
			\includegraphics[width=\textwidth]{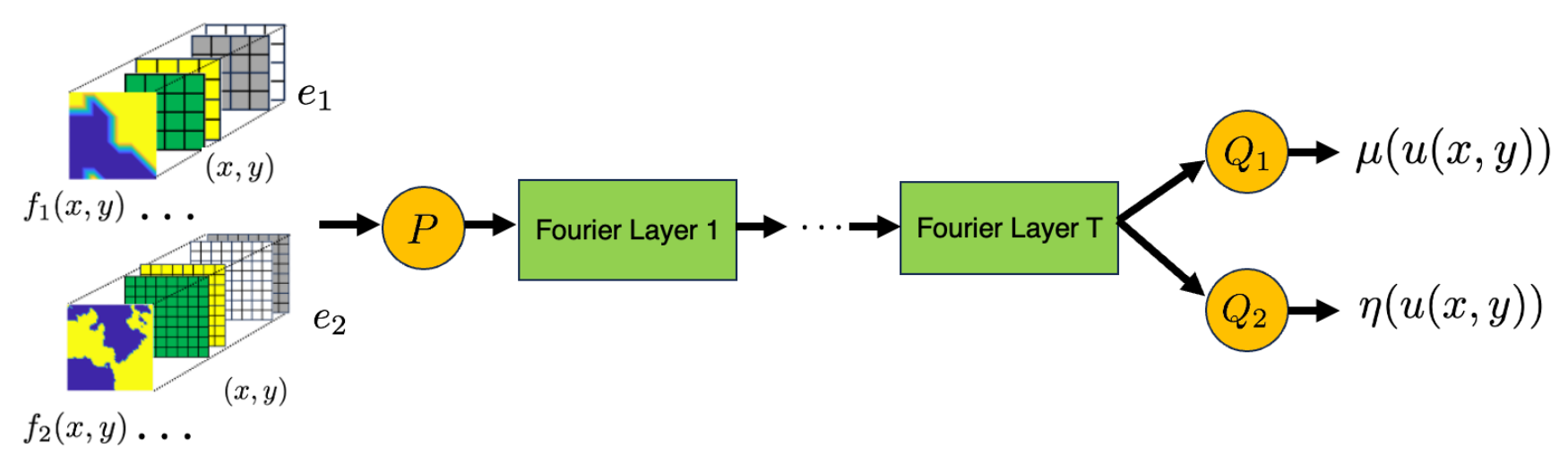}
		\end{subfigure}
	\end{tabular}
	%\vspace{-0.1in}
	\caption{\small Graphical representation of our probabilistic multi-resolution FNO. Here $P$ is the FFN that lifts the input function to a higher-dimensional channel space, $Q_1$ is the FFN for channel-wise projection and producing the prediction mean, 
		and $Q_2$ is a convolution net plus another FFN to produce the variance in the log space. } \label{fig:model}
\end{figure*}
\begin{figure*}[h!]
	\centering
	\setlength\tabcolsep{0pt}
	\begin{tabular}[c]{ccc}
		\setcounter{subfigure}{0}
		\begin{subfigure}[t]{0.33\textwidth}
			\centering
			\includegraphics[width=\textwidth]{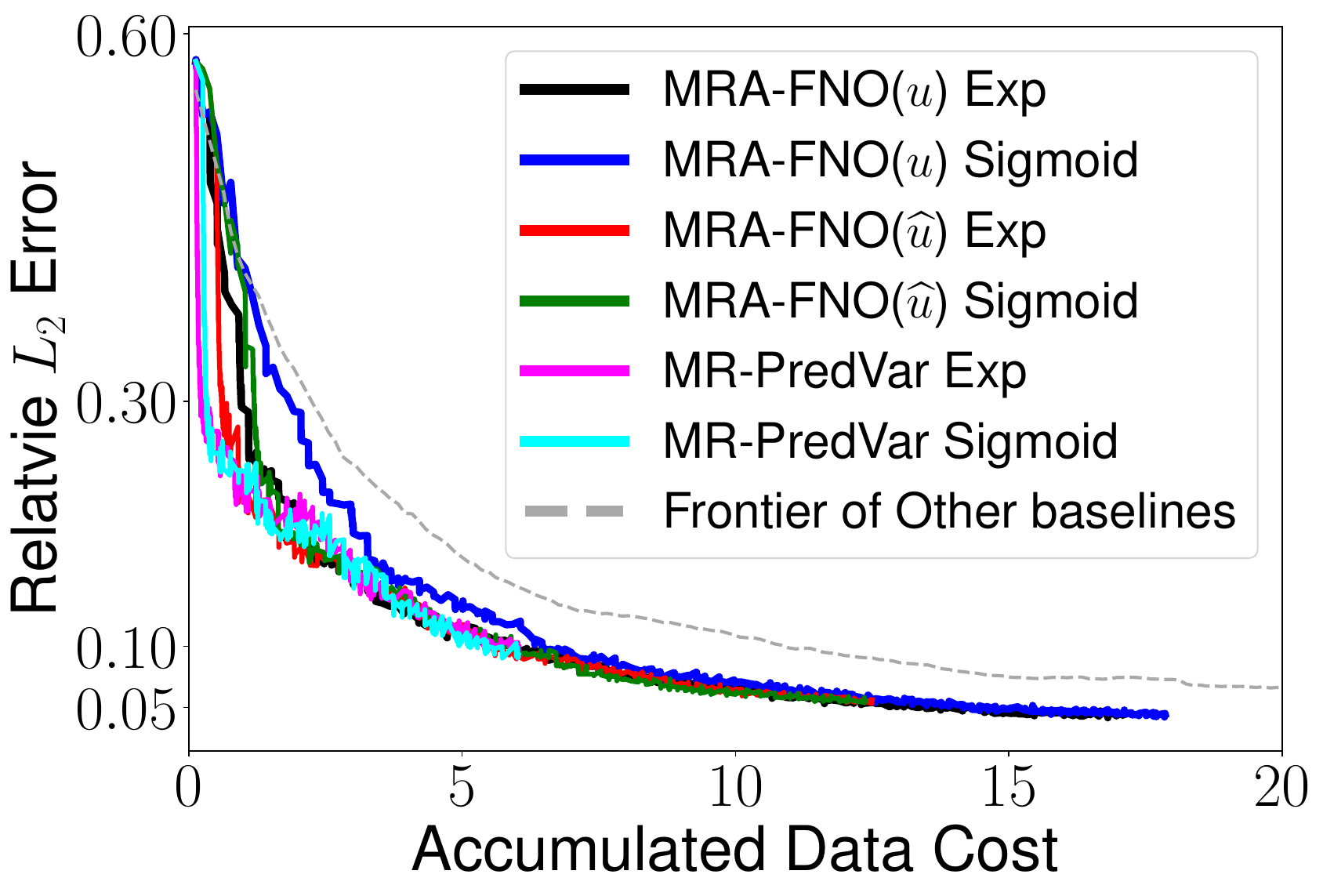}
			\caption{\small \textit{Burgers}}
		\end{subfigure}
		&
		\begin{subfigure}[t]{0.33\textwidth}
			\centering
			\includegraphics[width=\textwidth]{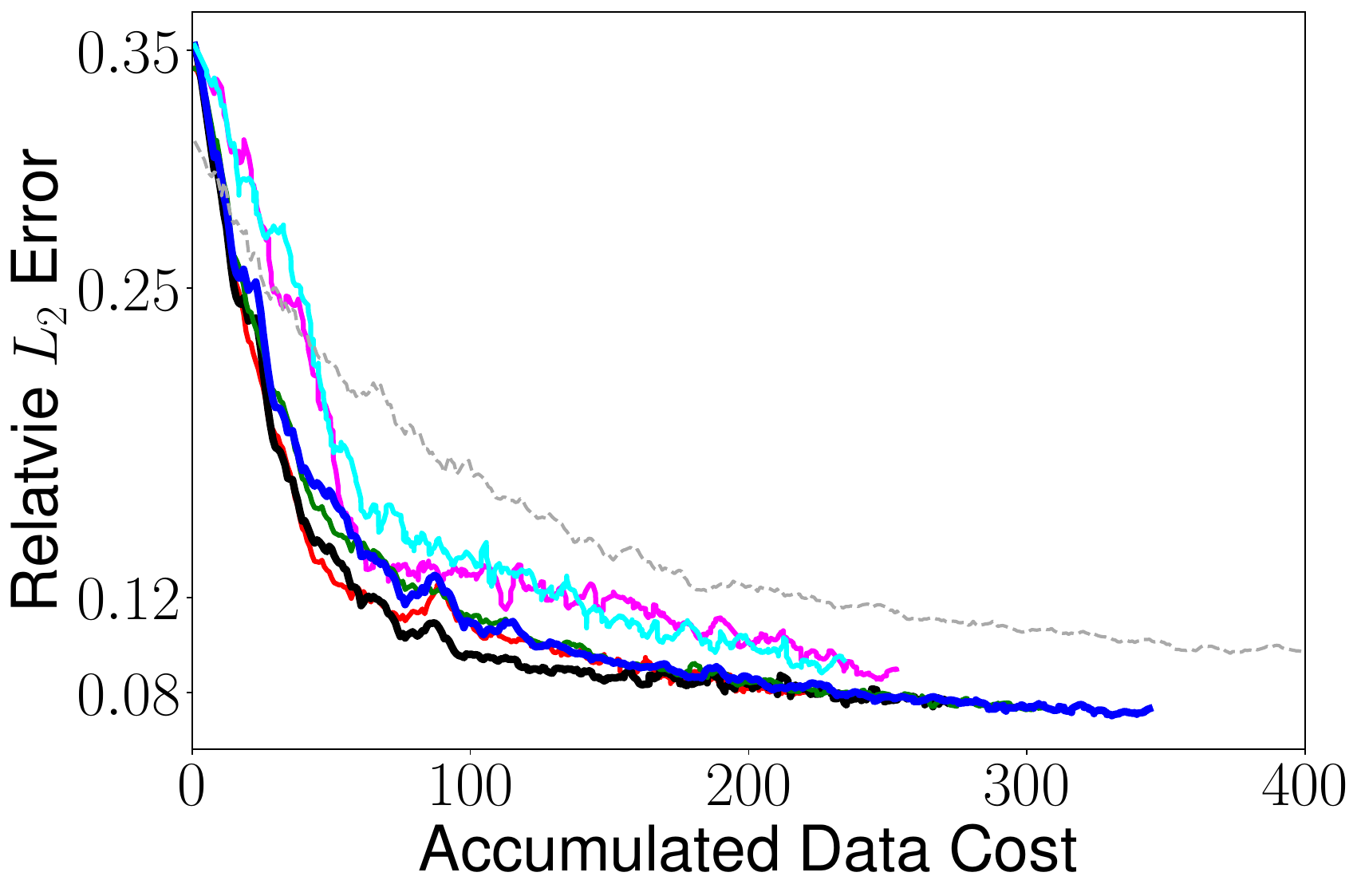}
			\caption{\small \textit{Darcy}}
		\end{subfigure}
		&
		\begin{subfigure}[t]{0.33\textwidth}
			\centering
			\includegraphics[width=\textwidth]{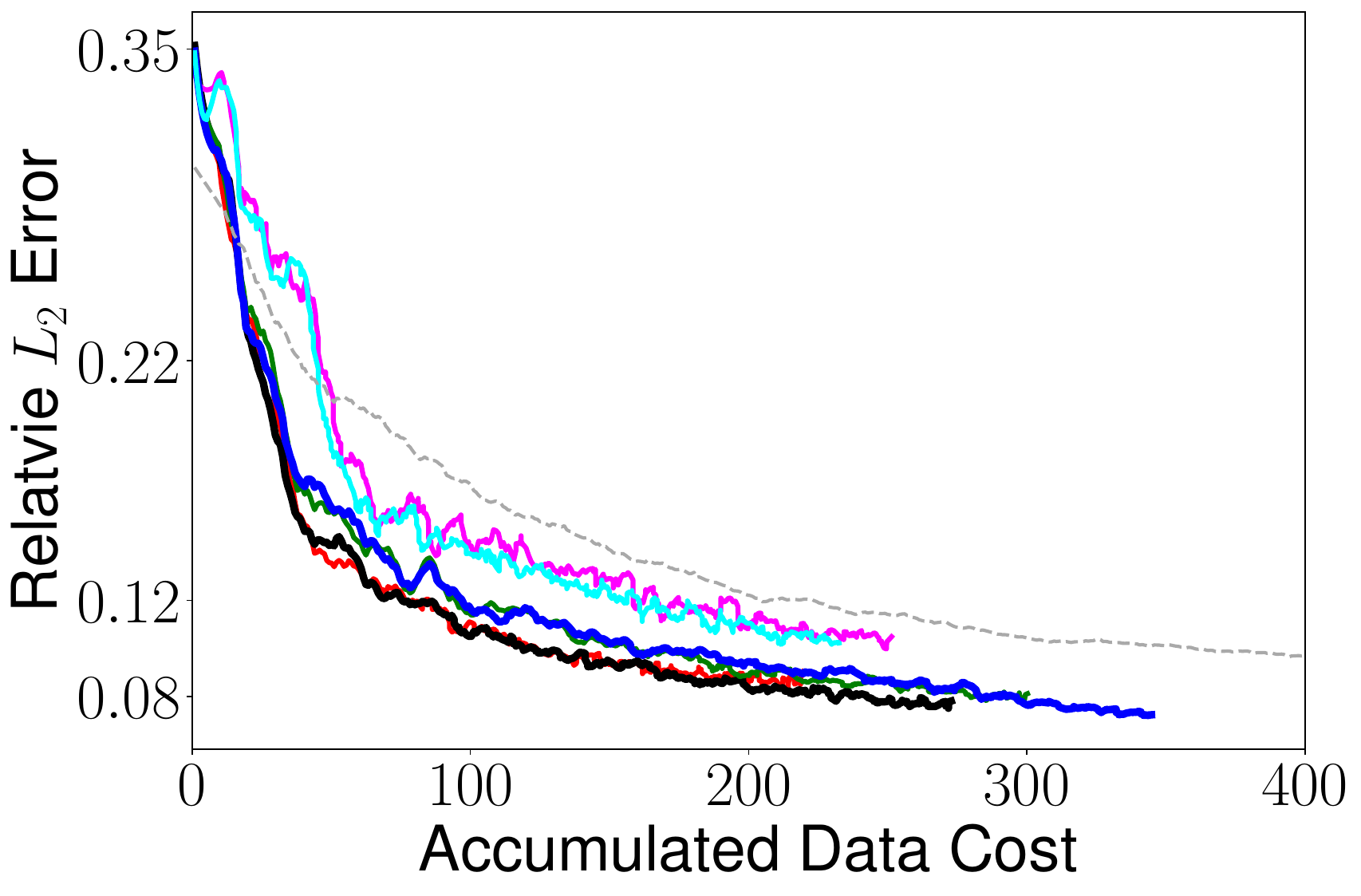}
			\caption{\small \textit{Darcy3}}
		\end{subfigure}\\
		\begin{subfigure}[t]{0.33\textwidth}
			\centering
			\includegraphics[width=\textwidth]{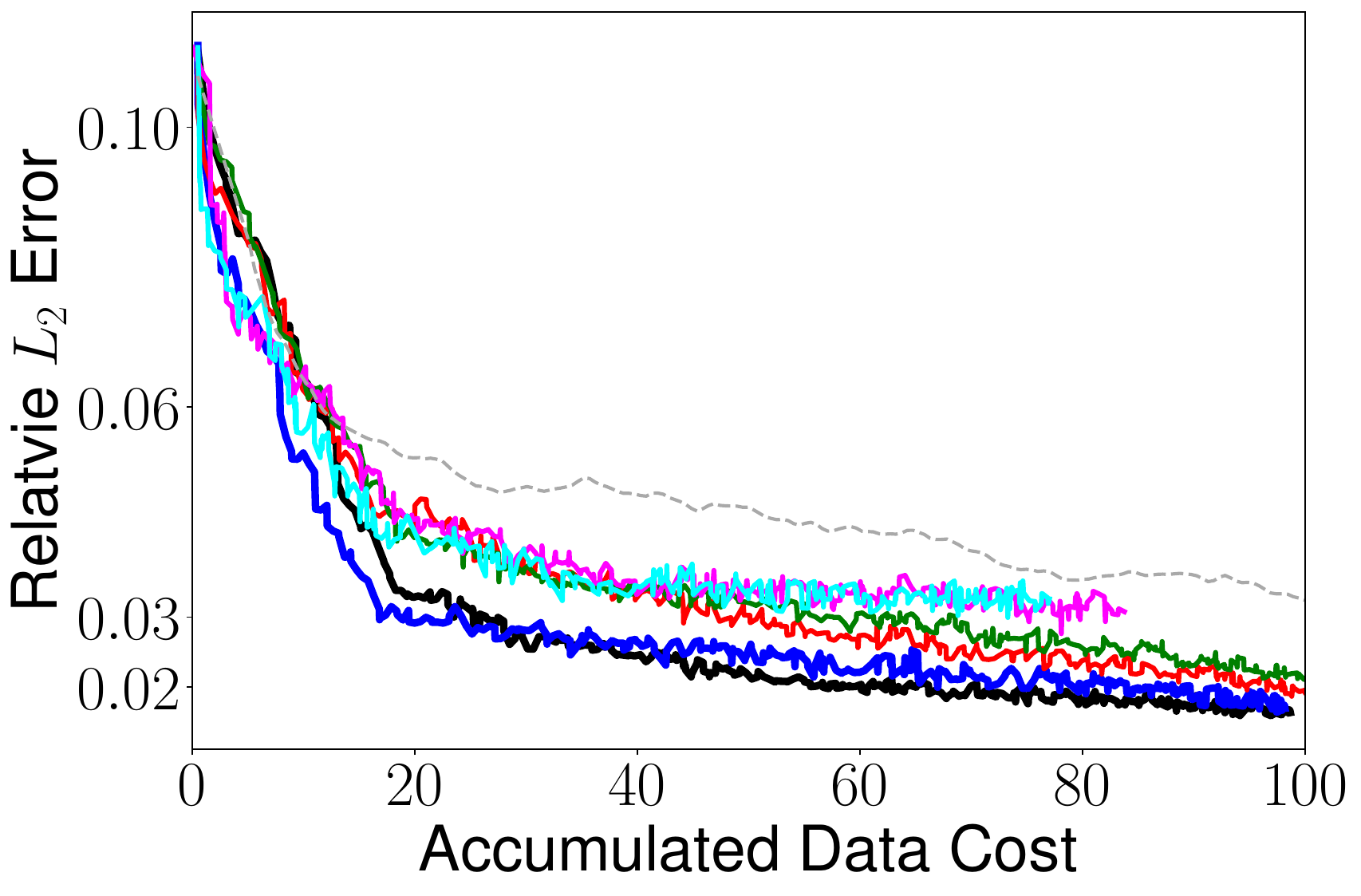}
			\caption{\small \textit{Diffusion}}
		\end{subfigure}
		&
		\begin{subfigure}[t]{0.33\textwidth}
			\centering
			\includegraphics[width=\textwidth]{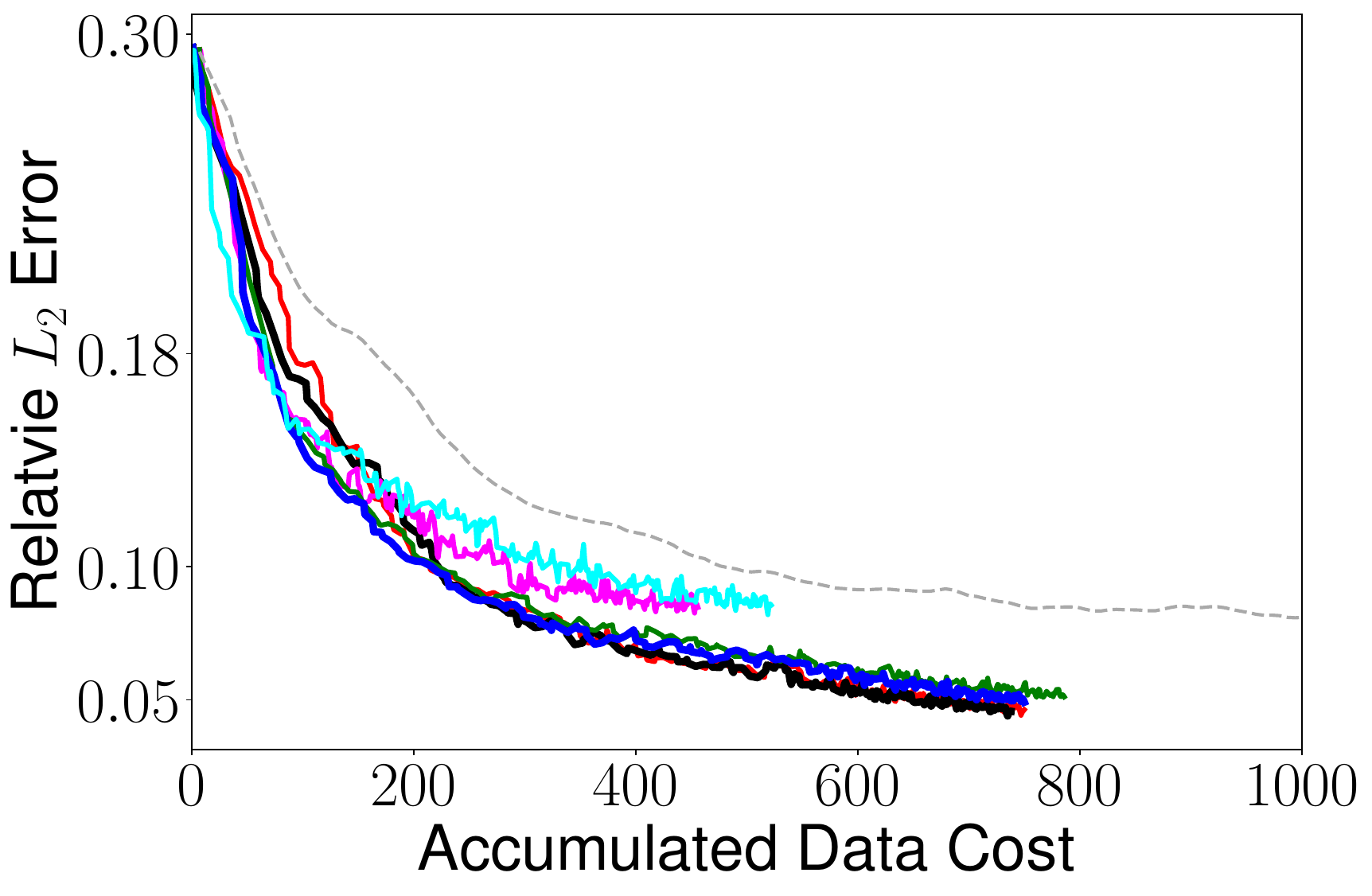}
			\caption{\small \textit{NS}}
		\end{subfigure}
		&
		\begin{subfigure}[t]{0.30\textwidth}
			\centering
			\includegraphics[width=\textwidth]{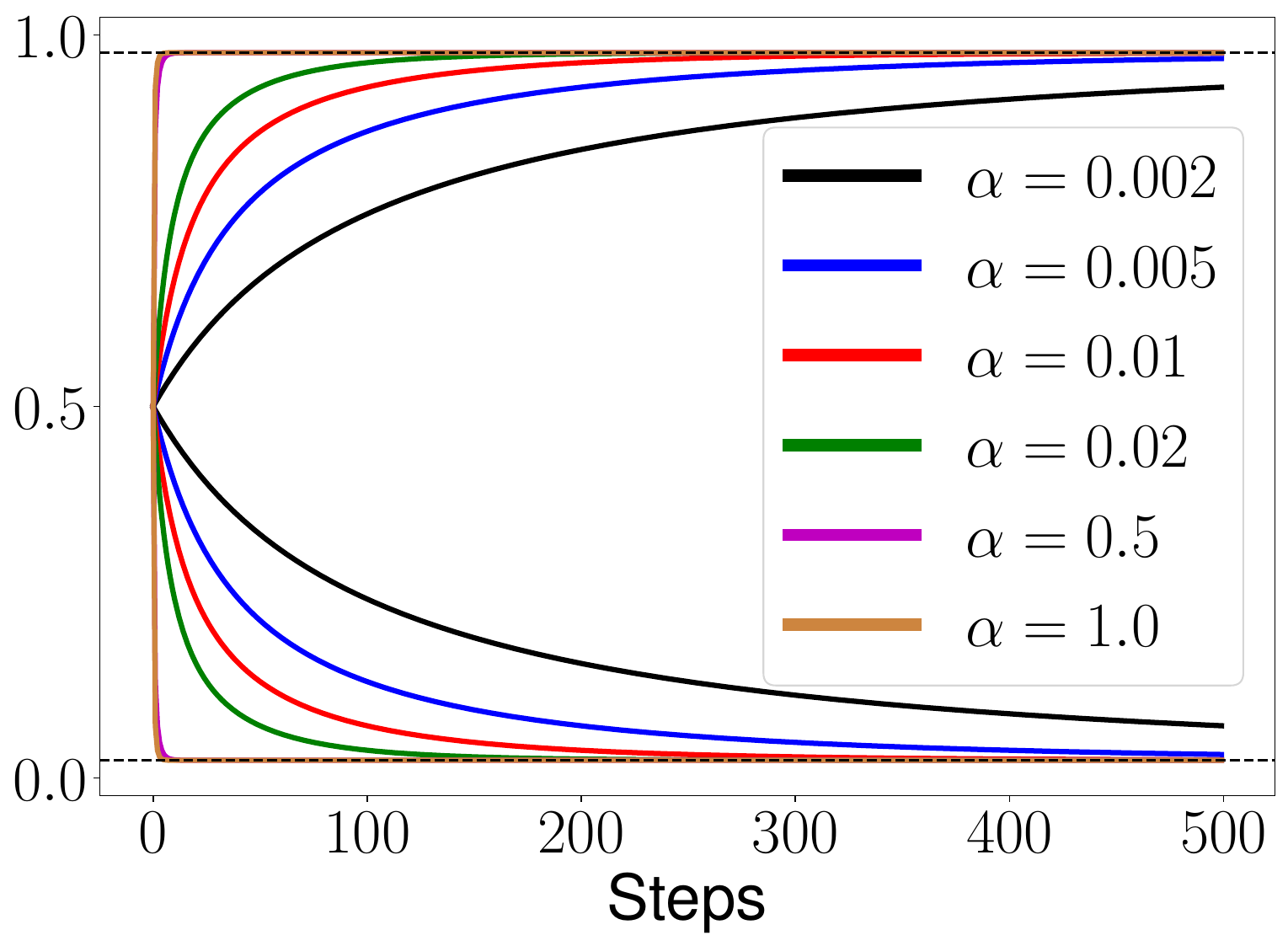}
			\caption{\small \textit{Cost Scheduler (Sigmoid)}}
		\end{subfigure}
	\end{tabular}
	\caption{\small Relative $L_2$ error \textit{vs.} accumulated data cost (a-e) and the cost schedule with a sigmoid-based decay.  Each method ran 300 active learning steps for NS, and 500 steps for all the other tasks.  Note that different methods can end up with different total data cost (after running the same number of steps).} \label{fig:acc-more}
	
\end{figure*}

\begin{figure*}
	\centering
	\setlength\tabcolsep{0pt}
	\includegraphics[width=0.6\textwidth]{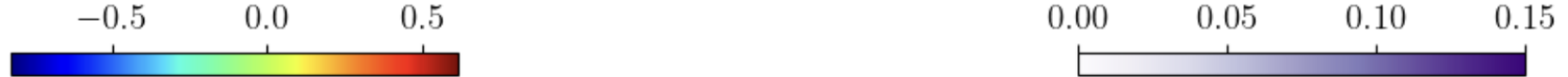}
	\begin{tabular}[c]{c}
		\setcounter{subfigure}{0}
		\begin{subfigure}[t]{1.0\textwidth}
			\centering
			\includegraphics[width=\textwidth]{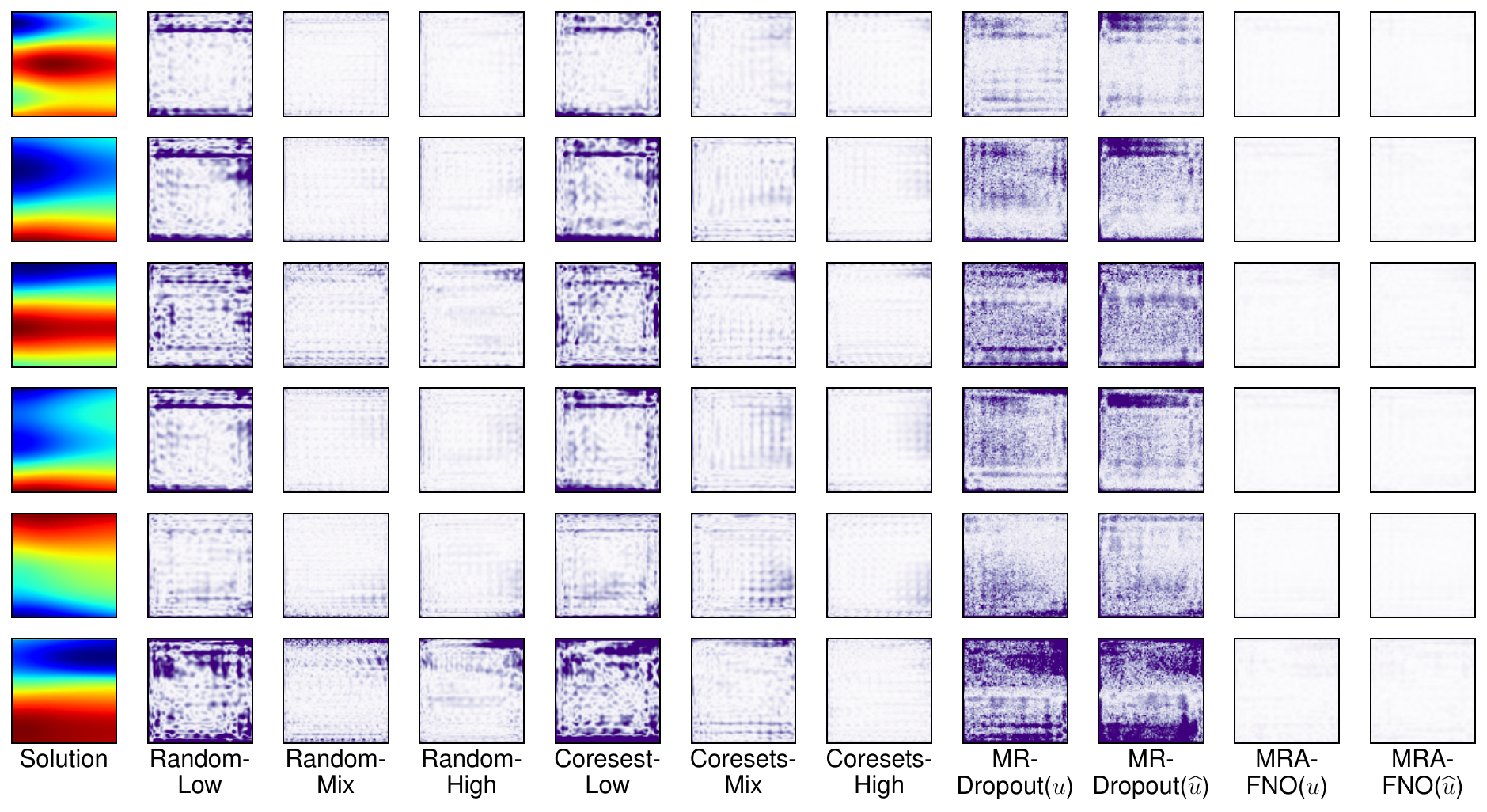}
		\end{subfigure}
	\end{tabular}
	%\vspace{-0.1in}
	\caption{\small Point-wise error on  nonlinear \textit{Diffusion}.} \label{fig:point-wise-diffusion}
\end{figure*}

\cmt{
	\begin{figure*}[t]
		\centering
		\setlength\tabcolsep{0pt}
		\includegraphics[width=0.6\textwidth]{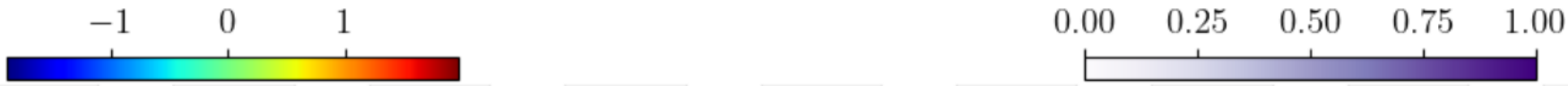}
		\begin{tabular}[c]{c}
			\setcounter{subfigure}{0}
			\begin{subfigure}[t]{1.0\textwidth}
				\centering
				\includegraphics[width=\textwidth]{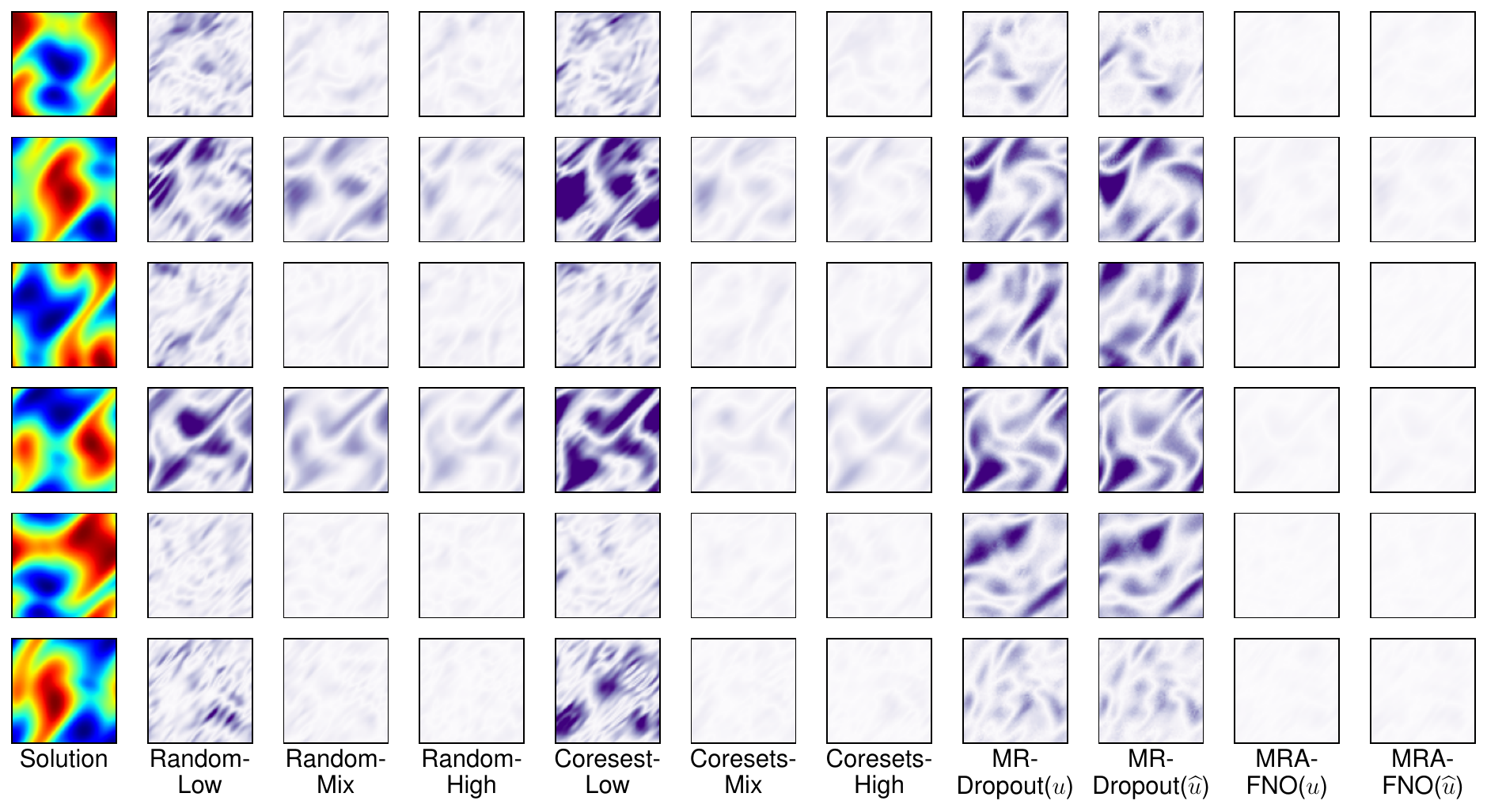}
			\end{subfigure}
		\end{tabular}
		%\vspace{-0.1in}
		\caption{\small Point-wise error on \textit{NS}.} 
		
	\end{figure*}
}